\documentclass[pdflatex,sn-nature]{sn-jnl}

\usepackage{graphicx}%
\usepackage{multirow}%
\usepackage[dvipsnames]{xcolor}
\usepackage{amsmath,amssymb,amsfonts}%
\usepackage{amsthm}%
\usepackage{mathrsfs}%
\usepackage{longtable}
\usepackage{booktabs}
\usepackage[title]{appendix}%
\usepackage{textcomp}%
\usepackage{manyfoot}%
\usepackage{booktabs}%
\usepackage{algorithm}%
\usepackage{algorithmicx}%
\usepackage{algpseudocode}%
\usepackage{listings}%
\usepackage{multirow}
\usepackage{siunitx}
\usepackage{comment}
\usepackage{caption}
\usepackage{subcaption}
\usepackage{hyperref}
\usepackage[table,xcdraw]{xcolor}  

\usepackage{booktabs,
            makecell, 
            tabularx}
\usepackage{textcomp}
\usepackage{mathtools}

\usepackage{tcolorbox}
\usepackage{xcolor}

\tcbuselibrary{theorems}
\tcbuselibrary{skins}
\tcbuselibrary{breakable}

\newtcolorbox{methodbox}{
    colback=blue!5!white,
    colframe=blue!75!black,
    fonttitle=\bfseries,
    breakable,
    enhanced
}

\newtcolorbox{clinicalbox}{
    colback=green!5!white,
    colframe=green!75!black,
    fonttitle=\bfseries,
    breakable,
    enhanced
}

\newtcolorbox{warningbox}{
    colback=red!5!white,
    colframe=red!75!black,
    fonttitle=\bfseries,
    breakable,
    enhanced
}

\newtcolorbox{conceptbox}{
    colback=orange!5!white,
    colframe=orange!75!black,
    fonttitle=\bfseries,
    breakable,
    enhanced
}

\theoremstyle{thmstyleone}%
\theoremstyle{thmstyletwo}%

\theoremstyle{thmstylethree}%

\begin{document}

\title[Article Title]{Causal Graph Neural Networks for Healthcare}


\author*[1]{\fnm{Munib} \sur{Mesinovic}}\email{munib.mesinovic@eng.ox.ac.uk}

\author[2, 3]{\fnm{Max} \sur{Buhlan}}\email{max.buhlan@medizin.uni-leipzig.de}

\author[1]{\fnm{Tingting} \sur{Zhu}}\email{tingting.zhu@eng.ox.ac.uk}

\affil*[1]{\orgdiv{Department of Engineering Science}, \orgname{University of Oxford}, \orgaddress{\city{Oxford}, \country{UK}}}

\affil[2]{\orgdiv{Faculty of Medicine}, \orgname{Leipzig University}, \orgaddress{\city{Leipzig}, \country{Germany}}}

\affil[3]{\orgdiv{Oxford Centre for Clinical Magnetic Resonance Research, Medical Sciences Division}, \orgname{University of Oxford}, \orgaddress{\city{Oxford}, \country{UK}}}


\abstract{

Healthcare artificial intelligence systems often degrade in performance when deployed across institutions, with documented performance drops and perpetuation of discriminatory patterns embedded in data. This brittleness comes, in part, from learning statistical associations rather than causal mechanisms. Causal graph neural networks address this by combining graph-based representations of biomedical data with causal inference to learn invariant mechanisms instead of just spurious correlations. This Perspective reviews the methodology of structural causal models, disentangled causal representation learning, and techniques for interventional prediction and counterfactual reasoning on graphs. We discuss applications across psychiatric diagnosis and brain network analysis, cancer subtyping with multi-omics causal integration, continuous physiological monitoring, and drug recommendations. These methods provide building blocks for patient-specific Causal Digital Twins that could support in silico clinical experimentation. Remaining challenges include computational costs that preclude real-time deployment, validation challenges that go beyond standard cross-validation, and the risk of causal-washing where methods adopt causal terminology without rigorous evidentiary support. We propose a tiered framework distinguishing causally-inspired architectures from causally-validated discoveries and outline future directions, including scalable causal discovery, multi-modal data integration, and regulatory pathways for these methods. Making practical Causal Digital Twins possible will require an honest assessment of what current methods deliver, sustained collaboration across disciplines, and validation standards that match the strength of the causal claims being made.
}

\maketitle

\section{Introduction}

In 2019, a widely deployed risk prediction algorithm for healthcare resource allocation underestimated the health needs for Black patients by relying on healthcare costs as a proxy \cite{Obermeyer_2019_Dissecting}. A diabetic retinopathy screening system with high-level laboratory performance failed when deployed in community clinics. The system struggled with a kind of distribution shift and rejected patient images as ungradable due to lighting and protocol settings that differed from the settings in which the system was developed \cite{beede2020human}. These challenges reveal a triple crisis of AI in healthcare where models can degrade under distribution shift, perpetuate discriminatory patterns embedded in historical data, and struggle to provide reliable mechanistic insight \cite{Futoma_2020_Myth, Zech_2018_Confounding, Obermeyer_2019_Dissecting, ghassemi2021false}. In a pneumonia risk prediction study, models learned to classify asthma patients as lower risk because asthmatic patients received more aggressive treatment in the training hospitals, a confounding pattern that could have led to dangerous triage decisions had the models been deployed clinically  \cite{Caruana_2015_Intelligible}. The question facing healthcare AI is no longer whether systems can achieve impressive accuracy in retrospective test sets, but whether they can survive clinical deployment across diverse institutions and maintain both effectiveness and equity.

The brittleness occurs from a mismatch between what machine learning optimises and what healthcare often requires. Clinical decisions are often a result of causal mechanisms. Will this drug help \textit{this} patient? Would this patient have developed complications under alternative treatment? Current models learn statistical associations that may reflect institutional practices, equipment characteristics, patient selection, or confounding variables instead of biology \cite{Castro_2020_Causality, Richens_2020_Improving}. The COVID-19 pandemic exposed these limitations in two ways: predictive models built specifically for COVID-19 diagnosis and prognosis suffered from methodological shortcomings and poor generalisability \cite{Wynants_2020_Prediction, Roberts_2021_Common}, while pre-existing clinical AI systems, such as the widely deployed sepsis-alerting model, had to be deactivated when pandemic-driven changes in patient demographics disrupted the statistical relationships on which they relied \cite{finlayson2021clinician}. The consequences go beyond accuracy; they can also lead to patient harm, as models that encode spurious correlations may recommend interventions that are effective in training environments but ineffective or harmful in deployment settings. Moreover, evidence shows that AI systems trained on historical data perpetuate or amplify existing healthcare disparities, and that purely statistical fairness metrics are insufficient to address discrimination \cite{subbaswamy2020development, Obermeyer_2019_Dissecting,Kilbertus_2017_Avoiding, Kusner_2017_Counterfactual, Plecko_2024_Causal_Fairness, AdamsPrassl_2023_Algorithmic}.

Randomised controlled trials remain the gold standard for estimating causal effects, but as they address population-average treatment effects under specific protocols, they struggle to capture complex multi-drug interactions at scale, and are infeasible for many clinically relevant questions due to ethical, temporal, or cost constraints. When randomised trials are infeasible, causal inference frameworks offer principled approaches to counterfactual reasoning from observational data, including at the patient level \cite{Prosperi_2020_Nmi}. Causal graph methods can extend this logic to the high-dimensional molecular and physiological networks characteristic of modern biomedicine.

Pearl's Causal Hierarchy distinguishes between correlation and causation, and organises reasoning into three levels of increasing inferential power \cite{Pearl_2009_Causality, Hernan_2020_Causal}. Level 1 (Association) addresses observational queries through conditional probabilities $P(Y|X)$, answering ``what is?'' questions via pattern recognition, where standard machine learning does reasonably well. Level 2 (Intervention) on the effects of actions, formalised using the do-operator $P(Y|\text{do}(X))$, addresses ``what if we do?'' questions that are important for treatment planning. Level 3 (Counterfactual) queries alternative outcomes under hypothetical conditions, answering ``what would have been?'' questions useful for personalised medicine and retrospective analysis. Healthcare would benefit from Levels 2 and 3, predicting treatment responses, comparing therapeutic strategies, and identifying causal disease mechanisms, but standard models operate at Level 1, identifying statistical patterns without understanding the causal structure \cite{Pearl_2009_Causality, bareinboim2022pearl, Prosperi_2020_Nmi}. The Causal Hierarchy Theorem shows that these levels form a strict hierarchy where information at higher levels cannot be derived from lower levels without additional causal assumptions \cite{bareinboim2022pearl, Pearl_2009_Causality}. This partially explains why purely associational machine learning models, regardless of architectural sophistication or the scale of training data, are unable to reliably answer interventional and counterfactual questions needed for more robust clinical decision-making.

Biomedical systems naturally form networks, molecular interactions, brain connectivity, metabolic pathways, protein complexes, and disease comorbidity patterns, making graph representations a helpful framework for encoding biological relationships \cite{Barabasi_2011_Network, Nicholson_2023_Constructing, sporns2012simple}. Graph neural networks (GNNs) build on traditional graph analysis to learn representations from graph-structured data through iterative message passing, in which nodes aggregate information from their neighbours through learnable neural transformations that capture both local motifs and global network properties \cite{Hamilton_2020_Graph, Battaglia_2018_Relational, Kipf_2017_Semi}. This learning paradigm has demonstrated substantial advantages over handcrafted features in biomedical applications, ranging from molecular property prediction on chemical graphs to neurological classification based on brain connectivity \cite{Duvenaud_2015_Convolutional, corso2024graph}. Standard GNNs, however, inherit the limitations of supervised learning in optimising predictive performance by exploiting statistical patterns in the training data, whether reflecting biological mechanisms or spurious correlations arising from confounding, selection bias, or batch effects, and treat all network edges equally regardless of the underlying causal structure \cite{Bevilacqua_2021_Size, Knyazev_2019_Understanding, wu2022discovering}.

Causal inference combined with graph neural networks could address these challenges. Recent theoretical advances in causal representation learning suggest plausible ways forward. By modelling causal structures within graph architectures, we could develop models that identify therapeutic targets, predict treatment responses under interventions never observed in training data, and discover disease mechanisms beyond spurious correlations \cite{Schoelkopf_2021_Toward, Locatello_2020_Commentary}. This is particularly timely given four developments. First, the accumulation of multi-modal biomedical data, from single-cell sequencing that captures large-scale gene expression measurements per cell to continuous physiological monitoring that generates large amounts of data with temporal dynamics, provides strong resolution for causal discovery. Datasets now increasingly capture interventional outcomes and temporal dependencies that were previously unavailable \cite{Regev_2017_Human, Topol_2019_High, conroy2023uk}. Second, regulatory frameworks emphasise explainable AI in healthcare, which requires not just predictions, but a mechanistic understanding that clinicians can verify against domain expertise and biological knowledge \cite{rudin2019stop, us2024transparency, Murdoch_2021_Definitions}. Third, previously mentioned weaknesses of correlation-based models under distribution shift cannot satisfy healthcare's need for robust generalisation through causal invariance. Fourth, merging causal inference with fairness constraints would help identify discrimination transmitted along specific causal pathways and help distinguish predictive factors from discriminatory proxies, central to anti-discrimination law, but absent from traditional machine learning approaches \cite{Plecko_2024_Causal_Fairness, Plecko_2025_Fairness_Accuracy}.

This Perspective examines causal graph neural networks, methods that integrate causal inference with graph-based deep learning, as a methodological framework for moving AI in healthcare from pattern recognition to proactive causal medicine. Unlike previous papers that addressed GNNs in biomedical data \cite{Li_2022_Graph} or causal inference in epidemiology \cite{Hernan_2020_Causal, Petersen_2021_Causal} separately, we focus on their integration, methods in which both graph structure and causal reasoning can achieve clinical impact. We analyse how causal principles enhance GNN architectures for healthcare tasks and examine current applications that demonstrate utility across psychiatric diagnosis, cancer treatment, physiological monitoring, and drug discovery. We also identify challenges in scaling these methods to real-world deployment, including computational complexity, validation without ground truth, fairness under distribution shift, and regulatory considerations. Our analysis shows both promise and substantial barriers. Although causal graph neural networks can address the challenges mentioned above, they face computational requirements that limit real-time application, difficulty validating causal claims from observational data, and gaps between methodological sophistication and clinical interpretability.

We propose the development of patient-specific \textit{Causal Digital Twins}, dynamic computational models built on causal graph neural network frameworks that allow clinicians to perform in silico experiments before clinical intervention. The digital twin paradigm emerged as a framework for personalised medicine, offering dynamic computational representations of individual patients that evolve through iterative data inclusion \cite{EmmertStreib_2025_DigitalTwins, Laubenbacher_2022_Roadmap}. Four capabilities can distinguish medical digital twins from conventional simulation models: \textit{explainability} through mechanistic rather than black-box architectures, \textit{intervenability} enabling virtual testing of therapeutic scenarios, \textit{learnability} allowing patient-specific model refinement over time, and \textit{diversability} supporting uncertainty quantification through ensemble approaches \cite{EmmertStreib_2025_DigitalTwins}. However, extending digital twins to complex multisystemic diseases requires methodological foundations that current data-driven approaches do not provide. Causal graph neural networks could provide the theoretical foundation. Structural causal models enable interventional reasoning for virtual treatment testing, counterfactual computation supports patient-specific outcome prediction, and invariant causal learning can potentially ensure robust generalisation across healthcare contexts. The integration of causal inference with graph-based patient modelling can thus enable us to transform digital twins from aspirations into methodologically grounded clinical tools. Imagine a clinician treating advanced cancer who could load the patient’s multi-omics profile, brain imaging, and clinical history into such a system, then simulate candidate treatment strategies to compare predicted outcomes. This vision, while ambitious, rests on methodological foundations now being established through causality-inspired architectures for interventional prediction, counterfactual generation, and robust multimodal integration. Alongside this, we need to rethink how we evaluate and deploy AI in medicine, shifting from predictive accuracy on retrospective data to causal validity in prospective deployment, from statistical fairness metrics to interventional equity, and from black-box pattern recognition to models whose reasoning clinicians can scrutinise against domain knowledge. This will require advances in scalable causal discovery algorithms, validation frameworks suited to healthcare constraints, integration strategies for multimodal, multiscale biological data, and regulatory pathways for systems that make causal claims. This rests on collaboration among computational scientists, clinicians, biomedical researchers, and policymakers, united by the recognition that healthcare AI’s next decade will be defined by which approaches achieve not just strong metric performance but also clinical trust through mechanistic understanding.

\begin{figure*}[!]
\centering
\includegraphics[width=\textwidth]{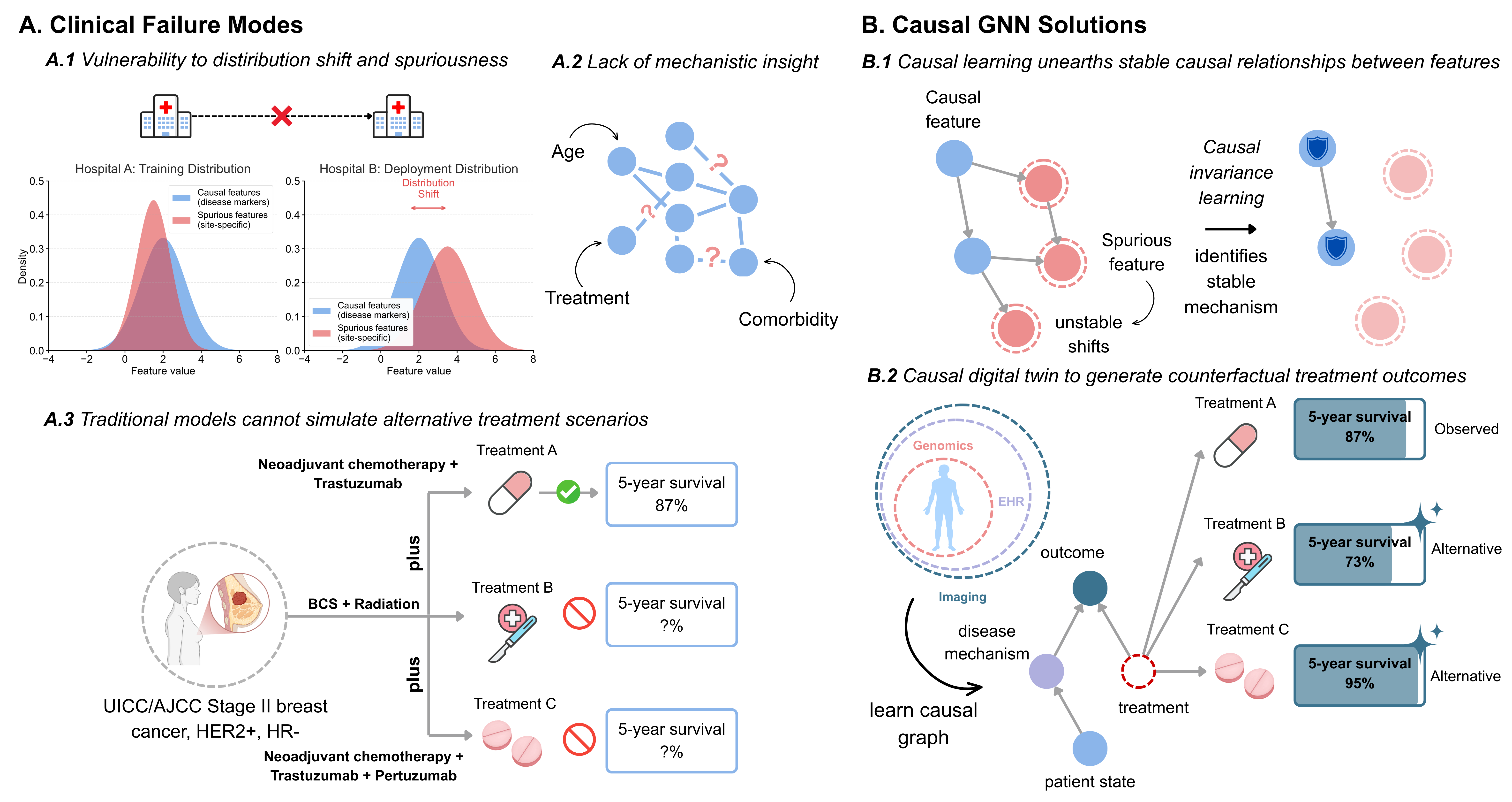}
\caption{Limitations of traditional machine learning and solutions using causal graph neural networks. 
\textbf{A}, Clinical failure modes illustrate limitations of correlation-based models in healthcare deployment. 
\textbf{A.1}, Vulnerability to distribution shift: A screening system achieves 94\% accuracy at the training institution (Hospital A), but performance collapses to 73\% at deployment sites (Hospital B), for example, due to spurious correlations with site-specific protocols and patient demographics rather than causal disease mechanisms. Feature distributions show that spurious features (red, site-specific) undergo a distributional shift, while causal features (blue, disease markers) remain invariant across environments. 
\textbf{A.2}, Lack of mechanistic insight: traditional models operate as black boxes, learning associations between age, treatment, and comorbidity without identifying the underlying causal pathways. Question marks indicate uncertain causal relationships that cannot be disentangled from observational data alone. 
\textbf{A.3}, Counterfactual blindness: Associational models trained on observational data (Treatment A, 87\% five-year survival) cannot simulate alternative treatment scenarios (Treatments B and C) marked with prohibition symbols, as they lack the structural causal knowledge required for personalised interventional reasoning.
\textbf{B}, Causal GNN offer some solutions through causal inference. 
\textbf{B.1}, Causal invariance learning: multi-environment optimisation finds stable causal relationships (blue nodes with shield protection) while suppressing environment-specific spurious correlations (red nodes, faded), ensuring robust generalisation under distribution shift through invariant risk minimisation. 
\textbf{B.2}, Causal digital twin for counterfactual generation: integration of multi-modal patient data (shown as concentric rings) constructs patient-specific structural causal models of disease mechanisms. Do-operator helps simulate unobserved treatment scenarios and generate counterfactual predictions (Treatment B: 73\%, Treatment C: 95\% five-year survival) through interventional inference without requiring empirical observations. BCS = Breast Conserving Surgery, HR-: Hormone receptor negative. This figure displays a subset of treatment options and does not represent the entire range of possible interventions \cite{LeitlinienprogrammOnkologie2025Mammakarzinom}. Icons were taken from BioRender.}
\label{fig:causal_failure_solutions}
\end{figure*}

Causal graph neural networks address limitations through principled causal inference frameworks (Fig.~\ref{fig:causal_failure_solutions}). These methods can be built towards identifying invariant biological mechanisms rather than spurious correlations, improve robustness in deployment across heterogeneous clinical settings and support mechanistic treatment optimisation through counterfactual reasoning. This Perspective proceeds in three stages. Section 2 establishes the theoretical foundations connecting structural causal models to graph neural network architectures. Section 3 presents methodological advances organised by clinical capability, disentanglement, intervention prediction, counterfactual generation, robustness, and fairness, providing the technical building blocks. Section 4 demonstrates clinical impact, by showing how these methods translate into diagnosis, prognosis, treatment selection, and monitoring across healthcare domains. Section 5 synthesises these advances toward Causal Digital Twins, articulating how the methodological components combine to enable patient-specific in silico experimentation. Section 6 addresses the substantial barriers, computational, validation, and translational, that must be overcome for clinical deployment. We note that Sections 2--4 primarily synthesise existing methods and empirical evidence, while Section 5 articulates an aspirational framework, Causal Digital Twins, whose full realisation remains prospective. Section 6 addresses the substantial methodological, computational, and translational barriers separating current capabilities from this vision.

\section{The causal graph framework}

\subsection{Why biological systems suit graph-structured causal models}

Biological systems exhibit causal relationships organised as networks across multiple scales of organisation. At the molecular level, gene regulatory networks encode transcriptional control in which regulatory proteins (causally) influence the expression of target genes, with modern single-cell RNA sequencing that captures large-scale expression profiles \cite{badia2023gene, Regev_2017_Human}. Protein-protein interaction networks represent physical binding relationships that mediate cellular signalling, with perturbations propagating through these networks to generate phenotypic responses \cite{Barabasi_2011_Network}. At the tissue level, brain networks comprise hundreds of regions with functional and structural connectivity, where activity in one region can drive or modulate activity in connected regions \cite{Schirmer_2021_Graph}. At the population level, disease comorbidity networks from large-scale data mining of electronic health records and genetic association data are integrated into knowledge graphs to link conditions through shared features \cite{Nicholson_2023_Constructing, jensen2014temporal}. This graph structure makes neural network architectures that represent and reason about network topology a well-suited framework for biomedical machine learning.

The mathematical background that connects causal inference to graph representations is based on Structural Causal Models (SCMs), a rigorous formalisation of cause-and-effect relationships \cite{Pearl_2009_Causality}. An SCM $\mathcal{M} := \langle U, V, F, P(U) \rangle$ specifies exogenous variables $U$ determined outside the model, endogenous variables $V$ defined within the model, structural equations $F = \{f_1, \ldots, f_m\}$ where each $V_i = f_i(\text{pa}(V_i), U_i)$ with $\text{pa}(V_i)$ denoting the parent set of $V_i$, comprising all variables that directly cause $V_i$ in the causal graph, and probability distribution $P(U)$ over exogenous variables \cite{bareinboim2022pearl, Pearl_2010_Introduction}. This induces a directed acyclic graph $\mathcal{G}(\mathcal{M})$ in which nodes represent variables and directed edges encode functional dependencies through the parent sets $\text{pa}(V_i)$. The joint distribution factorises according to the Markov property: $P(V_1, \ldots, V_m) = \prod_{i=1}^{m} P(V_i | \text{pa}(V_i))$ \cite{Pearl_2009_Causality}. Interventions in this framework are ``graph surgery'' (adapted from Pearl's mini-surgery term), modifying the causal graph by removing incoming edges to intervened variables and setting their values, thus altering the generative process \cite{Pearl_2009_Causality, Zecevic_2021_Relating}. This graph-theoretic representation of causation suggests that message-passing operations may serve as a natural computational support for causal inference.

Causal graphs are shaped by modelling choices on the semantic definition and resolution of nodes \cite{Chalupka_2017_Causal, Rubenstein_2017_Causal}. In biomedical applications, nodes may represent entities at vastly different levels of abstraction, such as individual genes versus pathways, single neurons versus brain regions, specific metabolites versus metabolic modules. This granularity is not just a technical convenience; it also determines the content of the causal relationships described and the interventions the graph implies \cite{Woodward_2003_Making}. The causal structure discovered at one resolution may not transfer to finer or coarser scales, as the relationships that persist between brain regions might decompose into opposing sub-regional effects, while molecular interactions may aggregate to emergent pathway-level causation invisible at the gene level \cite{hoel2013quantifying}. Throughout this Perspective, we tried to note the resolution at which different methods operate.

Different biological networks carry different causal weight. Gene regulatory networks nominally encode regulatory relationships: perturbing transcription factor X should alter the expression of target gene Y. However, GRNs inferred from observational expression data risk conflating correlation with causation, where edges reflect shared upstream regulators or cellular-state confounding rather than direct regulation. CausalBench, a benchmark using large-scale single-cell perturbation data, found that poor scalability limits the performance of existing network inference methodss \cite{Chevalley_2025_CausalBench}. An initial evaluation showed that established interventional methods did not outperform observational-only approaches, contrary to expectations based on synthetic benchmarks, although methods developed in a subsequent community challenge that more effectively leveraged interventional data achieved substantially higher performance. ODE-based systems biology models for metabolic and signalling dynamics sit on a different footing as they encode mechanistic hypotheses grounded in biochemical knowledge, such as enzyme kinetics and reaction stoichiometry, and so approximate causal mechanisms more directly, though parameters fitted to observational data may still capture environment-specific rather than invariant relationships. Protein-protein interaction networks, used by methods such as PDGrapher as proxy causal graphs \cite{Roohani_2024_PDGrapher}, mix direct physical interactions with indirect associations mediated through protein complexes, making their causal interpretation uncertain. We propose that physiological or biological networks are better treated as candidate causal structures that require validation through interventional evidence. Causality-informed and graph neural network
based (CiGNN) takes a step in this direction for the modelling of physiological signals by applying causal discovery algorithms to learn the relationships between the features of the wearable sensor data \cite{Liu_2024_CiGNN}.

Standard GNNs will learn from whatever statistical pattern best predicts the label, regardless of whether it reflects causal relationships. In resting-state brain imaging, physiological noise from cardiac and respiratory cycles inflates connectivity estimates across regions \cite{birn2012role}; in molecular networks, batch effects fabricate associations between proteins that never physically interact \cite{goh2017protein}; in clinical data, confounding by indication means drug-outcome correlations often trace back to prescribing behaviour rather than therapeutic effect \cite{kyriacou2016confounding}. These shortcuts predict well within the training distribution but degrade under the shifts encountered in deployment. Causality-inspired architectures try to separate the signal that holds across environments from the noise that does not. The Debiasing framework for GNNs via Disentangled Causal Substructure (DisC) uses a shared edge mask generator to split each graph into causal and bias subgraphs, and trains separate GNN modules with distinct loss functions so that the causal partition carries the primary predictive signal \cite{fan2022debiasing}. Causality-inspired graph neural network (CI-GNN) takes a different approach for brain networks, using a conditional mutual information constraint rooted in Granger causality to isolate the subgraph most relevant to the diagnostic label \cite{Zheng_2024_CIGNN}. Neither method delivers causal inference in the formal sense, but both push GNNs towards learning a structure that generalises, rather than a structure than one that merely fits statistically well.

\subsection{From association to intervention to counterfactuals}

Standard supervised learning performs well on associational queries, given covariates $X$, predicts the outcome $Y$ by learning $P(Y|X)$ from observational data. This is sufficient for many pattern recognition tasks: identifying malignant tumours from images, predicting patient deterioration from vital signs, or classifying disease subtypes from gene expression. However, clinical decision-making often requires interventional reasoning: what happens if we administer treatment $T$? This requires estimating $P(Y|\text{do}(T))$, which differs from $P(Y|T)$ whenever confounding exists, and confounding may pervade observational healthcare data through indication bias, selection effects, and unmeasured factors \cite{Hernan_2020_Causal, Pearl_2009_Causality}.

Graph-structured causal models support interventional prediction through Pearl's do-calculus, a set of three rules for converting interventional distributions into quantities estimable from observational data, given the structure of the causal graph (Box 1) \cite{Pearl_2009_Causality}. The first rule specifies when conditioning on additional observed variables can be dropped, based on d-separation in the manipulated graph. The second allows for replacing an intervention with an observation when the appropriate graphical independence condition is met. The third allows removing an intervention altogether when it has no causal effect on the outcome through the relevant paths. Zečević et al. \cite{Zecevic_2021_Relating} establish a theoretical connection between graph neural networks and structural causal models, and define a model class (the interventional VGAE) for which causal effect identifiability is equivalent to standard graph identifiability. The interventional VGAE (iVGAE) uses interventional GNN layers in the encoder and the decoder, thereby defining a model class that can, in principle, distinguish causal from associational structure in the learnt representations, though practical validation has been limited to synthetic settings.

Counterfactual reasoning is at the top of Pearl's hierarchy, addressing patient-specific questions such as whether a given patient would have developed complications under a different treatment \cite{Pearl_2009_Causality}. Computing counterfactuals involves abduction (updating beliefs about unobserved variables given evidence), action (modifying the causal model to reflect an alternative intervention), and prediction (deriving outcomes under the modified model) \cite{Pearl_2009_Causality}. In oncology, counterfactual models have been used to estimate individualised treatment effects over time, for example, to predict how different sequences of chemotherapy and radiotherapy would alter tumour trajectories for a specific patient \cite{bica2020estimating, bica2021real}. Separately, graph counterfactual methods such as the generative CounterfactuaL ExplAnation geneRator (CLEAR) generate counterfactual explanations on graphs using variational autoencoders, with an auxiliary variable mechanism that encourages the generated perturbations to respect underlying causal dependencies  \cite{Ma_2022_Clear}. Extending such graph-level counterfactual approaches to biomedical networks, for example, to explore how perturbations to molecular interaction graphs might reshape predicted disease outcomes, remains an open direction.

\begin{figure*}[!]
\centering
\includegraphics[width=\textwidth]{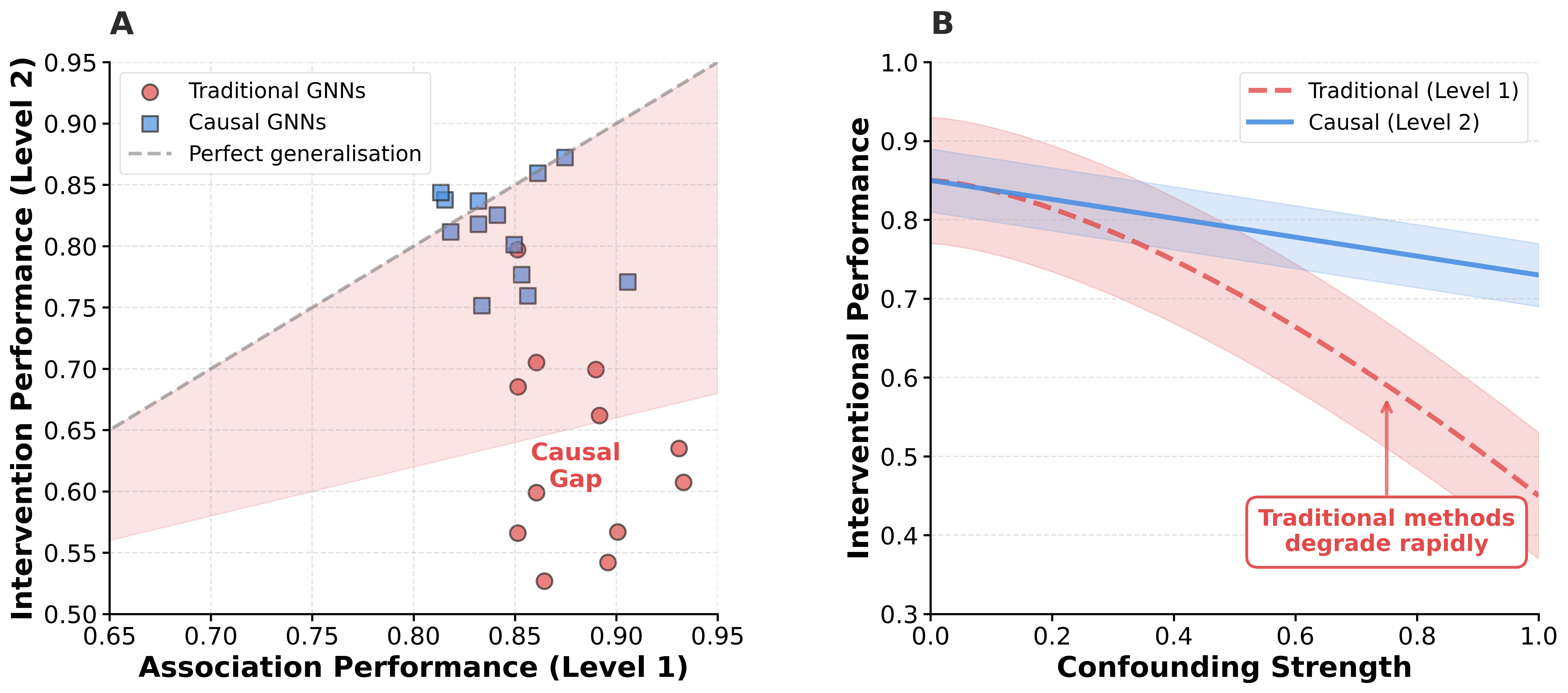}
\caption{Conceptual framework of theoretical advantages of causal over associational inference across Pearl's causal hierarchy.
\textbf{A}, Difference between associational and interventional performance predicted by causal hierarchy theory. The ``causal gap'' quantifies expected performance loss when models trained on observational data are applied to interventional queries, and associational patterns may not hold under intervention.
\textbf{B}, Robustness to simulated unmeasured confounding. Conceptual curves show the theoretical expectation that traditional methods degrade more steeply as unmeasured confounder strength increases, while causal architectures that model confounding structure are expected to maintain robustness. Confounding strength parameterised as the maximum risk ratio for unmeasured confounder associations with treatment and outcome. Shaded regions represent idealised uncertainty bands with expected variance patterns. \textit{Causally-Inspired GNN} (Tier 1) denotes architectures employing causal concepts, such as invariance regularisation to improve robustness without explicit causal claims. \textit{Causally-Validated GNN} (Tier 3) denotes methods whose causal claims are corroborated through multi-modal evidence triangulation. See Section 6.3 for definitions.}
\label{fig:causal_hierarchy}
\end{figure*}

Patient-specific \textit{Causal Digital Twins}, dynamic computational models that allow clinicians to perform in silico experiments, are the long-term goal. Such models would need to bring together a patient's multi-omics profile (genomic, transcriptomic, proteomic), longitudinal imaging of organ structure and function, clinical history including prior treatments and responses, and knowledge graphs encoding known biological mechanisms \cite{bjornsson2019digital, moingeon2023virtual}. A causal GNN architecture supporting such a twin could learn patient-specific parameterisations of general biological mechanisms and simulate therapeutic interventions using the do-operator framework, setting treatment variables, propagating effects through learnt causal pathways, and predicting downstream outcomes. In oncology, for example, a digital twin of this kind might evaluate candidate drug regimens by predicting their effects on disease-relevant pathways and estimating toxicity trade-offs, selecting personalised therapy before a single dose is given \cite{moingeon2023virtual}. In neurology, personalised virtual brain models have already been used to simulate seizure propagation and test surgical strategies in silico for patients with drug-resistant epilepsy \cite{jirsa2023personalised}; extending this paradigm to broader neuropsychiatric applications through causal graph architectures remains an open challenge. The methodological components include causal representation learning theory, which provides the formal machinery for interventional and counterfactual reasoning  \cite{Schoelkopf_2021_Toward}, while recent work has shown that deep architectures combining graph and convolutional models can recover causal structure from high-dimensional biomedical data under realistic noise and sample-size constraints \cite{lagemann2023deep}. Figure~\ref{fig:causal_hierarchy}A illustrates the theoretical basis and shows the conceptual hierarchy through which standard graph neural networks may match causal models on associational tasks but are expected, according to the Causal Hierarchy Theorem, to degrade on the interventional and counterfactual queries that clinical decision-making often requires.

Throughout this Perspective, we distinguish methods by the strength of their causal claims: \textit{causally-inspired} architectures that employ causal concepts without explicit causal claims, \textit{causally-structured} methods that estimate causal quantities under stated assumptions, and \textit{causally-validated} approaches whose claims are corroborated through multi-modal evidence (see Section 6.3 for definitions).

\begin{tcolorbox}[colback=blue!5!white,colframe=blue!75!black,title=\textbf{Box 1: Foundations of causal graph neural networks}]

\textbf{Structural Causal Models} \\
An SCM $\mathcal{M} := \langle U, V, F, P(U) \rangle$ defines the data-generating process, where exogenous variables $U = \{U_1, \ldots, U_n\}$ represent unmodelled background factors, endogenous variables $V = \{V_1, \ldots, V_m\}$ are defined by structural equations $V_i = f_i(\text{pa}(V_i), U_i)$, and $P(U)$ governs exogenous randomness \cite{Pearl_2009_Causality}. The induced causal graph $\mathcal{G}(\mathcal{M})$ contains directed edges $V_j \to V_i$ for each $V_j \in \text{pa}(V_i)$.

\textbf{Interventions and the do-operator} \\
Intervention $\text{do}(X = x)$ modifies the SCM by replacing the structural equation for $X$ with $X := x$, removing incoming edges in the graph. The post-intervention distribution $P(Y|\text{do}(X = x))$ differs from observational $P(Y|X = x)$ whenever confounding exists \cite{Pearl_2009_Causality}.

\textbf{Graph neural network implementation} \\
GNNs implement message-passing: $\mathbf{h}_i^{(l+1)} = \sigma\left(\mathbf{W}^{(l)} \mathbf{h}_i^{(l)} + \sum_{j \in \mathcal{N}(i)} \mathbf{M}^{(l)} \mathbf{h}_j^{(l)}\right)$, where $\mathbf{h}_i^{(l)}$ is the representation of node $i$ at layer $l$, $\mathcal{N}(i)$ are neighbours, and $\mathbf{W}, \mathbf{M}$ are learnable parameters. This architecture naturally aligns with causal graphs when edges represent causal relationships and message-passing implements causal influence propagation \cite{Zecevic_2021_Relating}.

\textbf{Causal identifiability}  \\
$P(Y|\text{do}(X))$ is identifiable if it can be expressed in terms of the observational distribution $P(V)$ given the causal graph $\mathcal{G}$. The do-calculus provides three rules for such expressions \cite{Pearl_2009_Causality}:
\begin{align}
P(Y|\text{do}(X), Z, W) &= P(Y|\text{do}(X), W) \text{ if } Y \perp_{\mathcal{G}_{\overline{X}}} Z | X, W \label{eq:rule1}\\
P(Y|\text{do}(X), \text{do}(Z), W) &= P(Y|\text{do}(X), Z, W) \text{ if } Y \perp_{\mathcal{G}_{\overline{X}\underline{Z}}} Z | X, W \label{eq:rule2}\\
P(Y|\text{do}(X), \text{do}(Z), W) &= P(Y|\text{do}(X), W) \text{ if } Y \perp_{\mathcal{G}_{\overline{X}\overline{Z(W)}}} Z | X, W \label{eq:rule3}
\end{align}
where $\mathcal{G}_{\overline{X}}$ denotes the graph with edges into $X$ removed, $\mathcal{G}_{\underline{Z}}$ has edges out of $Z$ removed, and $\perp$ indicates d-separation. Rule 1 permits ignoring observations on variables that do not affect the outcome through paths independent of the intervention; Rule 2 allows converting interventions to observations when appropriate conditional independence holds; Rule 3 enables removing interventions when they do not influence the outcome through certain graph paths.

\textbf{Counterfactual computation} \\
For unit $u$ and intervention $X = x$, the counterfactual outcome $Y_x(u)$ is computed via: (1) Abduction: infer $U = u$ from observed data; (2) Action: modify SCM to $\mathcal{M}_x$ with $X := x$; (3) Prediction: evaluate $Y$ under $\mathcal{M}_x$ with $U = u$ \cite{Pearl_2010_Introduction}. This enables patient-specific predictions unavailable from population-level interventional distributions.

\end{tcolorbox}

\section{Methodological advances}

We organise methodological advances by the clinical capabilities they enable: disentangling causal mechanisms from confounding, predicting interventional effects without experimental data, generating patient-specific counterfactuals, achieving robust generalisation across heterogeneous settings, and ensuring fairness through causal criteria. Within each capability, we discuss the technical approaches.  \textit{Disentanglement methods} (Section 3.1) decompose observed graphs into causal and spurious components through architectural separation, sometimes using multi-environment data. \textit{Structural causal methods} (Sections 3.2, 3.3) estimate interventional or counterfactual quantities using do-calculus or causal discovery algorithms, requiring stronger assumptions, including graph specification and causal sufficiency. \textit{Invariant learning methods} (Section 3.4) enforce feature stability across environments, assuming causal relationships hold across contexts while spurious correlations shift. \textit{Causal fairness methods} (Section 3.5) decompose outcome disparities along specific causal pathways, requiring a correctly specified causal graph relating sensitive attributes to outcomes. \textit{Knowledge-guided methods} incorporate biomedical priors, protein-protein interactions, gene regulatory networks, and anatomical atlases, as structural constraints without formal causal identification; these leverage domain expertise but inherit the epistemic status of their knowledge sources, which may conflate correlation with causation (Section 2.1). Table~\ref{tab:assumptions} maps individual methods to the relevant assumptions within this taxonomy.

\begin{table}[htbp]
\centering
\small
\caption{Causal assumptions and evidentiary status of methodological classes. Methods marked with * were developed outside clinical settings but are included for methodological relevance.}
\label{tab:assumptions}
\begin{tabular}{@{}p{0.14\textwidth}p{0.18\textwidth}p{0.32\textwidth}p{0.12\textwidth}p{0.14\textwidth}@{}}
\toprule
\textbf{Method Class} & \textbf{Representative Methods} & \textbf{Assumptions} & \textbf{Causal Claim} & \textbf{Tier} \\
\midrule
Disentanglement and invariant learning & DisC* \cite{fan2022debiasing}, CI-GNN \cite{Zheng_2024_CIGNN}, IGCL-GNN* \cite{Liu_2024_IGCLGNN}, DIR* \cite{wu2022discovering} & Observed graphs admit decomposition into causal and non-causal components; causal components remain predictive under distributional shift & Causal vs.\ spurious feature separation & Tier 1--2 \\
\addlinespace
Interventional prediction & Ze\v{c}evi\'{c} et al.* \cite{Zecevic_2021_Relating}, RC-Explainer* \cite{Wang_2022_RCExplainer}, CAL* \cite{Sui2022causal}, PDGrapher* \cite{Roohani_2024_PDGrapher},  CXGNN \cite{Behnam_2025_ECCV, behnam2024causal} & Graph structure is assumed known or approximated from domain knowledge; methods vary in whether confounders are explicitly modelled & Interventional reasoning or adjustment for confounding & Tier 1--2 \\
\addlinespace
Counterfactual generation & CLEAR* \cite{Ma_2022_Clear}, Huang et al.\ \cite{huang2025local} & Latent causal structure is at least partially identifiable from observational data; structural equations admit a learnable parameterisation & Counterfactual or causal explanation generation & Tier 1--2\\
\addlinespace
Causal discovery & CiGNN \cite{Liu_2024_CiGNN}, CautionGCN \cite{Wang_2024_CautionGCN} & Statistical dependencies reflect underlying causal structure (faithfulness); methods vary in whether latent confounders are permitted & Causal structure or feature identification & Tier 1--2 \\
\addlinespace
Fairness constraints & Plecko et al.\ \cite{plecko2025algorithmic}, Zhang et al.\ \cite{zhang2024causal}, path-specific methods \cite{Plecko_2024_Causal_Fairness} & Causal graph relating sensitive attributes to outcomes is correctly specified; relevant confounding pathways are measured & Path-specific effect decomposition & Tier 2 \\
\bottomrule
\end{tabular}
\begin{tablenotes}
\small
\item Tier 1: causally-inspired; Tier 2: causally-structured; Tier 3: causally-validated. See Section 6.3 for full tier definitions.
\item Assumptions are stated at the method-class level; individual methods may impose stronger or weaker versions.
\end{tablenotes}
\end{table}

\subsection{Disentangling causal signal from confounding}

Biomedical datasets contain spurious correlations that standard machine learning models, including GNNs, will happily exploit. In brain imaging, fluctuations in global arousal generate widespread correlations across regions that might not be causally linked to the neural processes under study \cite{gu2019arousal}. GNNs trained on such connectivity graphs can learn these confounds as if they were disease-relevant signals \cite{Zheng_2024_CIGNN}. In proteomic studies, batch effects introduce artificial associations between unrelated proteins and corrupt downstream analyses based on these measurements \cite{Cuklina_2021_BatchEffects}. In clinical records, drug co-occurrence patterns in prescribing data can produce correlations between molecular features and outcomes that reflect statistical regularities rather than therapeutic relationships, and recommendation systems trained on such data inherit these confounds unless corrected \cite{kyriacou2016confounding, Zhang_2023_CRec}. A common way to address these challenges is to decompose observed graphs into causal and spurious components. Fan et al. do this through disentangled causal substructure learning, using parameterised edge masks to separate causal from bias subgraphs and decorrelating their representations using counterfactual generation \cite{fan2022debiasing}. The DisC framework splits these input graphs into causal and bias subgraphs, each of which is processed by a dedicated GNN encoder \cite{fan2022debiasing}. Causal and bias representations are then decorrelated through counterfactual augmentation, in which bias latent vectors are randomly permuted across training graphs to synthesise unbiased samples. The edge masks learnt by DisC transfer across GNN architectures. Graphs pruned using masks from one base model improved generalisation when used to train structurally different GNNs, suggesting that the causal-spurious partition captures graph-level properties rather than artefacts of a particular encoder \cite{fan2022debiasing}. CI-GNN adapts similar principles to brain network analysis \cite{Zheng_2024_CIGNN}. CI-GNN learns disentangled subgraph-level representations under a graph variational autoencoder, using a conditional mutual information constraint inspired by Granger causality to separate causal from spurious connectivity. On the REST-meta-MDD dataset, standard GNN explainers identified widespread cerebellar hyperconnectivity in depressed patients. The causal decomposition of CI-GNN's gave far sparser cerebellar connections, suggesting that the hyperconnectivity detected by other methods reflected potentially spurious correlations rather than a disease-relevant signal. On the ABIDE dataset for autism spectrum disorder, CI-GNN identified tight connections within the somatomotor network, consistent with previous reports that somatomotor activation patterns characterise ASD brain states. In leave-one-site-out cross-validation across 17 ABIDE sites, CI-GNN's classification accuracy exceeded that of a comparison method at most sites, suggesting some robustness to inter-site heterogeneity. These results suggest that causal disentanglement can recover connectivity patterns more aligned with existing neurobiology.

\begin{figure*}[!t]
\centering
\includegraphics[width=\textwidth]{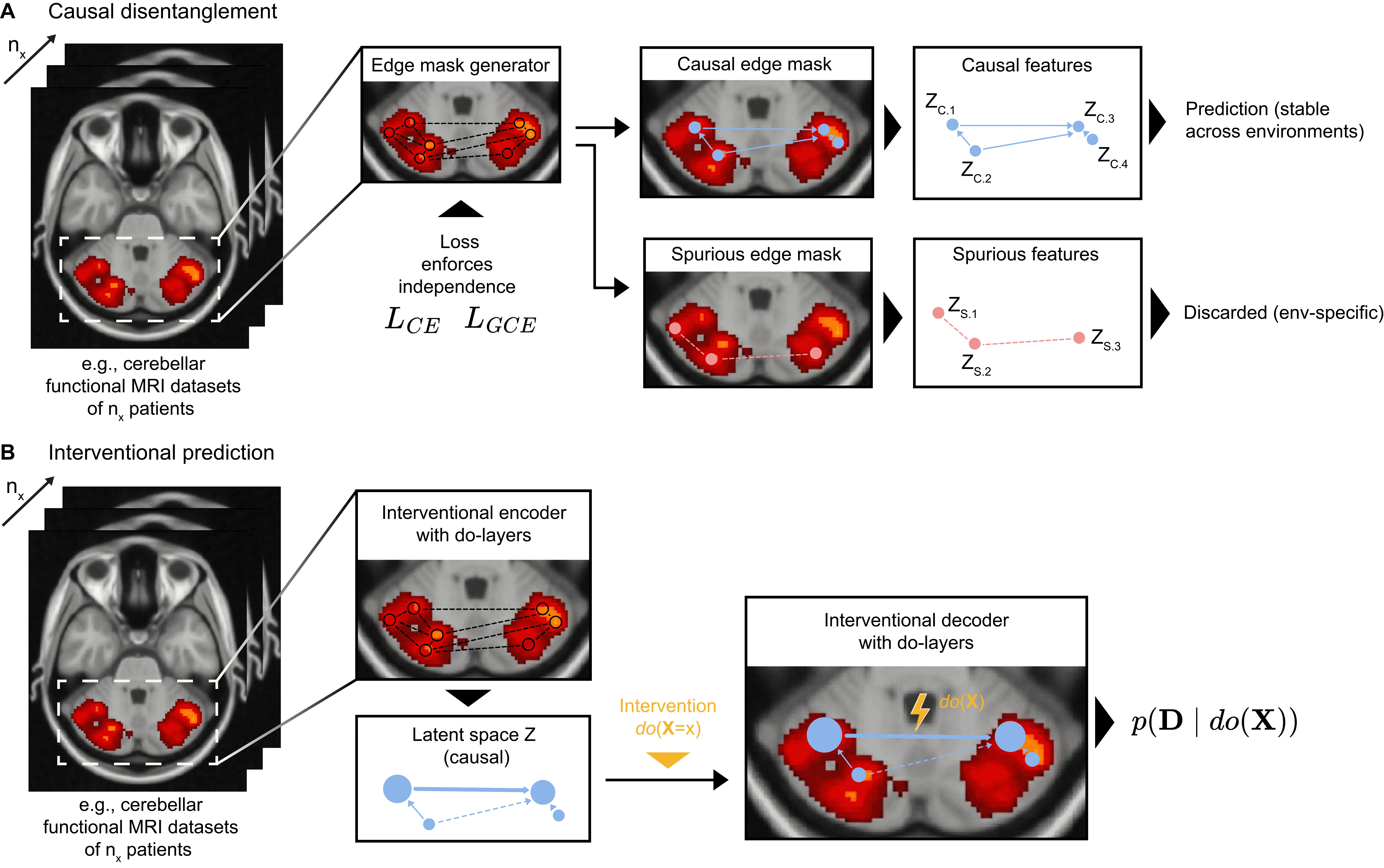}
\caption{Causal graph neural network principles illustrated for conceptual brain connectivity analysis. The underlying methods were developed on synthetic benchmark datasets; their application to functional neuroimaging is demonstrated to show their translational potential for clinical contexts.
\textbf{A}, Adapted from DisC (Debiasing via Disentangled Causal Substructure) framework \cite{fan2022debiasing} achieves causal disentanglement through a parameterised edge mask generator that partitions input graphs into causal and spurious subgraphs, each processed by a dedicated GNN encoder. The loss has two components: $\mathcal{L}_{CE}$, a weighted cross-entropy loss applied to the causal encoder that upweights samples the bias encoder misclassifies, directing the causal encoder toward features that predict outcomes independently of environmental shortcuts; and $\mathcal{L}_{GCE}$, a generalised cross-entropy loss that up-weights the gradient of the standard cross entropy loss for the samples with a high confidence, directing the bias encoder toward environment-specific shortcuts. Causal features ($Z_{C.1}$--$Z_{C.4}$) are decorrelated from spurious features ($Z_{S.1}$--$Z_{S.3}$) through counterfactual augmentation, in which bias representations are randomly permuted across training graphs.
\textbf{B}, Conceptual illustration of interventional reasoning on brain graphs, grounded in the framework of Ze\v{c}evi\'{c} et al.\ \cite{Zecevic_2021_Relating}. The figure depicts how such an architecture could map functional brain connectivity to a causal latent space and then apply Pearl's do-operator, $P(\mathbf{D}|\textit{do}(\mathbf{X}=x))$, to predict downstream effects of intervening on a specific brain region. Applied to brain networks, this could distinguish neural communication pathways from correlations induced by global arousal or shared confounders. The fMRI insets use the MNI152 template (Montreal Neurological Institute) distributed with FSL (FMRIB Software Library, University of Oxford) \cite{mazziotta1995probabilistic, jenkinson2012fsl}. The overlaid connectivity network is a schematic illustration by the authors and does not represent real patient data.}
\label{fig:causal_architectures}
\end{figure*}

CI-GNN operates on the AAL atlas, so the causal relationships it recovers describe functional circuit organisation at a larger scale resolution rather than individual neuronal connectivity \cite{Zheng_2024_CIGNN}. Figure~\ref{fig:causal_architectures} shows how two causal GNN architectures could in principle be applied to functional neuroimaging data. The DisC framework \cite{fan2022debiasing} addresses causal-spurious disentanglement through edge masking and counterfactual augmentation, while the theoretical results of Zečević et al. \cite{Zecevic_2021_Relating} establish that GNN architectures can in principle represent Pearl's do-operator, paving the way to interventional queries on graph-structured data.

\subsection{Predicting interventional effects without experimental data}

Precision medicine cannot rely solely on prediction; it must include a mechanistic explanation, because decisions need to account for the features driving a diagnosis. Standard GNN explainers rely on gradient or attention scores to highlight important edges, but these approaches often treat edges as independent and cannot separate causal substructures from spurious correlations present in the training data \cite{Wang_2022_RCExplainer, Sui2022causal}. RCExplainer addresses the first problem by using a reinforcement learning agent that sequentially adds edges to a candidate explanation, with each action rewarded based on its causal effect on the GNN’s prediction \cite{Wang_2022_RCExplainer}. Because the reward accounts for dependencies between the newly added edge and those already selected, the method captures coalition effects that independent attribution scores miss. Causal Attention Learning (CAL) treats shortcut features as confounders that open a backdoor path between causal subgraph features and the label, then applies the backdoor adjustment from causal theory to block this path \cite{Sui2022causal}. On synthetic graph benchmarks with controlled distribution shift, CAL maintained stable performance across bias levels.

For blood pressure monitoring, the CiGNN framework demonstrates how causal discovery can guide the design of the GNN architecture \cite{Liu_2024_CiGNN, Liu_2023_Causal}. Starting from 222 wearable features extracted from ECG and PPG signals, CiGNN applies fast causal inference (FCI) to infer an initial causal graph, then uses causal generative neural networks to orient the remaining ambiguous edges. The resulting graph identifies a reduced set of features causally linked to blood pressure variation, including PPG morphological features beyond the commonly used pulse transit time. A spatiotemporal GNN then takes this causal graph as input and exploits both the discovered causal structure and beat-to-beat temporal dynamics. Across subjects in normotensive and hypertensive groups, CiGNN outperformed comparison methods and remained accurate under perturbations that induce acute blood pressure changes \cite{Liu_2024_CiGNN}.

Zečević et al. provide theoretical grounding for this integration by proving that GNNs can subsume structural causal models \cite{Zecevic_2021_Relating}. Their framework formalises how to implement do-operator interventions within a GNN variational autoencoder, establishing that causal effects identifiable from the graph and observational data are also neurally identifiable within this model class, and vice versa. PDGrapher translates this kind of reasoning into drug discovery \cite{Roohani_2024_PDGrapher}. Given a pathological gene expression profile, PDGrapher predicts which combinatorial set of gene targets, if perturbed, would shift the cell towards a treated state. It embeds disease and treatment signatures into protein-protein interaction networks from BIOGRID, treating these as approximate causal graphs. Across cancer cell lines using chemical perturbation data, PDGrapher ranked ground-truth therapeutic targets significantly higher than the next-best method, and remained robust even as PPI network confidence thresholds were varied \cite{Roohani_2024_PDGrapher}.

\subsection{Generating counterfactuals for personalised medicine}

Counterfactual reasoning at the top of Pearl's causal hierarchy asks what would have happened under alternative conditions \cite{Pearl_2009_Causality}. In clinical settings, this translates to patient-level questions: Would a different treatment have changed the outcomes that are central to precision medicine and retrospective decision analysis \cite{Prosperi_2020_Nmi}. The CLEAR (Counterfactual Explanation Generator) framework tackles counterfactual generation on graphs using a graph variational autoencoder that maps input graphs to a latent space from which counterfactual graphs can be decoded \cite{Ma_2022_Clear}. To encourage the generated counterfactuals to respect causal dependencies rather than perturbing features independently, CLEAR introduces an auxiliary-variable mechanism inspired by nonlinear independent component analysis, enabling it to better identify the latent causal model without requiring explicit prior knowledge of the causal structure. The framework was validated on synthetic and real-world graph datasets, demonstrating improvements in counterfactual validity, proximity, and causal consistency over existing methods \cite{Ma_2022_Clear}.

A practical challenge for counterfactual reasoning on graphs is that simulated interventions can push graph structures well outside the training distribution, where model predictions become unreliable. Removing a highly connected node, for example, may produce a configuration that does not usually occur naturally in the observed data. In neuroscience, dynamic graph models are beginning to capture how brain connectivity evolves over time. The Dynamic Instance-level Graph Learning Network (DIGLN) learns patient-specific graph structures from intracranial EEG signals for seizure prediction, modelling evolving functional connectivity across spectral components of recording channels \cite{lian2025dynamic}. Recent work has begun to bring counterfactual reasoning into this setting, combining structural-functional brain coupling patterns with counterfactual explanation models to identify which connectivity changes would alter epilepsy predictions \cite{huang2025local}. Together, these directions point toward dynamic brain graph models that could support in silico testing of targeted interventions, though substantial methodological gaps remain before such models can reliably inform neurosurgical planning.

The GNN Causal Explainer (CXGNN) framework provides causal explanations for GNN predictions by building structural causal models over graph structures and parameterising the resulting structural equations as neural causal models \cite{Behnam_2025_ECCV}. This approach goes further than association-based GNN explainers, which are prone to spurious correlations, by computing cause-and-effect relationships among nodes to identify genuinely causal subgraphs. Applied to mild cognitive impairment (MCI) progression, Behnam et al. used this causal explanation approach on longitudinal clinical data, identifying hypertension, arrhythmia, congestive heart failure, coronary artery disease, stroke, lipid-related conditions, and sex among the variables forming the causal subgraph most informative for predicting MCI transitions \cite{behnam2024causal}. These findings align with evidence showing that vascular risk factors have sex-differentiated effects on cognitive decline, with hypertension and metabolic comorbidities bringing greater risk in women, while the burden of cardiovascular disease more adversely affects men \cite{gannon2019sex}. Such explanations of causal graphs could inform the counterfactual reasoning needed for personalised clinical models and identify which modifiable factors, if intervened upon, would alter a patient's predicted disease trajectory \cite{sanchez2022causal}.

\subsection{Achieving robustness across heterogeneous clinical settings}

Although disentanglement methods (Section 3.1) help robustness by separating causal from spurious representations, invariant learning approaches pursue the same goal through a different mechanism. They use optimisation constraints that enforce feature stability across environments without requiring architectural separation. Disentanglement offers an interpretable decomposition of causal and spurious factors, and invariant learning can give more portable models with simpler architectures, although its theoretical guarantees often rely on assumptions that may not hold in practice.

Deploying AI across healthcare institutions entails navigating differences in patient demographics, clinical protocols, and measurement technologies. These are the conditions under which models trained on statistical correlations tend to degrade. Deep learning models can tend to learn to exploit site-specific confounders rather than disease-relevant features \cite{Zech_2018_Confounding}. Invariant learning posits that causal relationships remain stable across contexts, whereas spurious correlations shift with the environment \cite{peters2016causal}. In clinical terms, the biological mechanisms of a treatment's efficacy should hold across populations, even as confounding factors, demographics, comorbidity profiles, and local prescribing habits vary (although pharmacogenomic differences can modulate these mechanisms at the individual level). Features that remain predictive across diverse settings are therefore more likely to reflect causal structure with a practical criterion for separating signal from noise in mind.

Methods such as the Information-based Gradient enhanced Causal Learning Graph Neural Network (IGCLGNN) demonstrate how multi-objective optimisation can balance the enhancement of causal features with the suppression of non-causal shortcut patterns in graph representations, although this framework has not yet been tested in clinical settings \cite{Liu_2024_IGCLGNN}. The underlying principle that robust deployment requires balancing predictive performance on current data with generalisation to unseen environments transfers to healthcare. As mentioned, in brain network analysis, CI-GNN uses Granger causality-inspired disentanglement to separate subgraph representations that are causally linked to diagnostic outcomes from those reflecting non-causal confounds, and has been evaluated on large multi-site psychiatric datasets spanning multiple sites \cite{Zheng_2024_CIGNN}. Although their primary contribution is interpretability, such causal decompositions are a natural precursor to cross-site robustness, in which features that determine a diagnosis should remain informative regardless of which scanner or protocol generated the data.

Moving from diagnosis to treatment, the Causal Representation learner for drug recommendation (CRec) framework applies similar causal disentanglement principles to drug recommendation \cite{Zhang_2023_CRec}. In observational prescribing data, drug co-occurrence patterns can create misleading associations, for example, proton pump inhibitors may appear to correlate with better cardiovascular outcomes simply because they are frequently co-prescribed with aspirin for gastric protection, not because of any direct cardiac benefit. CRec separates drug molecular representations into causal and spurious components at the atom level, conditioned on each patient's health state, then applies causal intervention to eliminate the spurious signal. Evaluated on the MIMIC-III and MIMIC-IV datasets, the framework showed improved recommendation accuracy over methods that do not account for such confounding \cite{Zhang_2023_CRec}. At the graph level, the DIR framework constructs interventional distributions by perturbing non-causal subgraph components rather than relying on predefined environment labels, and then identifies the subgraph rationales that remain predictive across these synthetic distributional shifts  \cite{wu2022discovering}. This provides an invariant learning strategy applicable to any graph classification setting, including molecular and clinical graphs, without requiring explicit knowledge of what the confounding environments are.

\subsection{Ensuring causal fairness in clinical algorithms}

Traditional fairness metrics based on statistical parity operate at Pearl's Level 1 (association), equalising predictions across demographic groups without considering how those predictions were generated~\cite{Pearl_2009_Causality, Plecko_2024_Causal_Fairness}. However, we argue that some clinical deployment decisions require at least Level 2 (intervention) reasoning. The question that matters is not just whether predictions correlate equally across demographics, but whether they remain stable under hypothetical interventions on sensitive attributes. A model can satisfy statistical parity while leaving the discriminatory causal mechanisms in the data-generating process intact.

Interventional fairness requires that the predicted outcome distributions remain invariant when the sensitive attribute is set by intervention: $P(\hat{Y} = y \mid \text{do}(A = a)) = P(\hat{Y} = y \mid \text{do}(A = a'))$ for all outcomes $y$ and attribute values $a, a'$ \cite{Plecko_2024_Causal_Fairness}. This extends to path-specific constraints. In the Standard Fairness Model, with sensitive attribute $A$, confounders $Z$, mediators $W$, and outcome $Y$, the total variation decomposes into counterfactual direct, indirect, and spurious effects, each transmitted along a distinct causal pathway \cite{Plecko_2024_Causal_Fairness}. A causally $S$-fair predictor constrains a chosen set $S$ of these pathways to transmit no discriminatory signal while preserving the causal structure along pathways outside $S$ \cite{Plecko_2025_Fairness_Accuracy}. Plecko et al.\ used this causal fairness decomposition to investigate racial and ethnic disparities in ICU outcomes between Indigenous and non-Indigenous patients in Australia and between African-American and White patients in the United States \cite{plecko2025algorithmic}. The mortality disparity pointed in opposite directions in the two cohorts, but the causal decomposition revealed consistent underlying mechanisms across both countries. Along the indirect pathway, minority patients faced higher mortality risk driven by worse chronic health, greater illness severity, and a higher proportion of urgent admissions. The direct effect of minority status was protective, however, when all other variables were held equal, minority patients showed improved survival, attributed to a selection effect in which poorer access to primary care leads to ICU admission for less severe, potentially preventable conditions \cite{plecko2025algorithmic}. Separately, Zhang et al.\ applied a causal fairness algorithm to EHR-derived coronary artery disease cohorts and found that women and Black patients were less likely to receive coronary artery bypass grafting than their counterparts \cite{zhang2024causal}. These results show how standard observational metrics can mask allocation disparities, or even reverse their apparent direction, that become visible only through causal decomposition.

Path-specific decomposition also helps clarify the cost structure of different fairness interventions~\cite{Plecko_2025_Fairness_Accuracy}. Removing direct effects of sensitive attributes on predictions often incurs minimal accuracy loss while substantially reducing disparity. Constraining indirect effects through mediators such as education or occupation imposes higher costs, reflecting predictive information carried along those pathways. This quantitative transparency would allow regulatory bodies, clinicians, and patient groups to weigh specific fairness-utility trade-offs based on causal evidence rather than post hoc statistical adjustments. A complementary finding reinforces the importance of problem formulation over post hoc correction. Obermeyer et al.\ showed that replacing a cost-based prediction label with a combined health-cost index reduced racial bias by 84\% without altering the model architecture or its input features, demonstrating that when the prediction target itself encodes structural inequality, no amount of algorithmic debiasing can substitute for choosing what to predict \cite{Obermeyer_2019_Dissecting}. For clinical deployment, such causal constraints offer the further advantage of enabling predictors to learn invariant causal relationships and are better positioned to maintain their fairness properties across healthcare contexts with different patient populations~\cite{Kilbertus_2017_Avoiding, Kusner_2017_Counterfactual}.

\section{Clinical impact across domains}

The translation of causal graph methods into clinical practice represents a convergence of methodological innovation with pressing healthcare needs. These applications span from individual patient diagnosis to population-level disease surveillance, demonstrating how causal inference transforms raw biomedical data into actionable clinical insights. Success depends not only on technical sophistication, but also on the addressing of fundamental clinical questions about disease mechanisms, treatment responses, and intervention strategies.

\subsection{Diagnosis: from correlation to mechanism}

\subsubsection{Mechanistic brain network analysis}

Brain network analysis shows why causal methods matter in neuroimaging. Standard graph-theoretical approaches have been useful, but tend to miss the high-dimensional, time-varying structure of brain connectivity \cite{Mohammadi_2025_BrainSci}. The brain can be mapped onto a graph, with regions of interest as nodes and functional connectivity as edges, yet correlation-based analyses often detect connections driven by global brain states, arousal shifts, or measurement artefacts rather than neural communication  \cite{Zheng_2024_CIGNN}.

CI-GNN learns disentangled latent representations that separate the causal factor behind a diagnostic subgraph from non-causal noise \cite{Zheng_2024_CIGNN}. It uses conditional mutual information, framed in terms of Granger causality, to quantify the causal influence of latent graph representations, from which a causal subgraph generator produces edge-level explanations that separate causally relevant connectivity from spurious correlations. Applied to major depressive disorder, CI-GNN flagged connections around the left rectus gyrus as prominent, while the cerebellum showed comparatively sparse connectivity \cite{Zheng_2024_CIGNN}. The clinical significance of these particular regional patterns remains open, but the shift from purely associative to causally grounded connectivity analysis is a substantive methodological advance that, however, needs to be complemented by multimodal neuroimaging \cite{cabalo2025multimodal} or electrophysiological approaches \cite{scholvinck2013contribution}.

Brain Prototype Interpretable Graph Neural Network (BPI-GNN) adds prototype learning to identify subtypes of psychiatric disorders from brain network data \cite{Zheng_2024_BPI}. Beyond separating patients from healthy controls, the model found subtypes with distinct symptom profiles. In depression, these differed on clinical subscales covering suicide risk, psychomotor retardation, and somatic symptoms \cite{Zheng_2024_BPI}. BPI-GNN outperformed existing brain network classifiers across multiple datasets. For epileptic seizure prediction, DIGLN learns patient-specific, time-varying graph structures that capture evolving functional relationships across spectral-spatial nodes derived from intracranial EEG channels \cite{lian2025dynamic}. By jointly modelling network organisation and temporal dynamics, DIGLN produced strong results on the Freiburg iEEG dataset, though external validation on larger and more varied cohorts is still needed.

\subsubsection{Multi-omics cancer subtyping}

Cancer classification depends on the integration of multiple data modalities. Single-omic approaches often lack the precision needed to establish robust associations between molecular-level changes and phenotypic traits \cite{Valous_2024_BrJCancer}. Tumours exhibit genomic instability, epigenetic dysregulation, transcriptomic reprogramming, and proteomic alterations, each providing complementary information about cancer biology \cite{negrini2010genomic, zhang2024tumor}.

Causal methodology offers an alternative to standard differential expression analysis for cancer subtyping. Rather than using all differentially expressed features, the CautionGCN framework applies a causal multi-head autoencoder to select genes with inferred causal relationships to cancer subtypes \cite{Wang_2024_CautionGCN}. Thousands of genes are differentially expressed across subtypes, but only a subset are plausible drivers. Like other causal discovery approaches, such methods typically operate under assumptions of causal sufficiency and faithfulness (Table~\ref{tab:assumptions}), which constrain the identifiability of the learnt graph structure. In addition to causal approaches, the Multi-Omics Graph cOnvolutional NETworks (MOGONET) framework uses graph convolutional networks for omics-specific learning and a View Correlation Discovery Network (VCDN) to explore cross-omics correlations in label space \cite{Wang_2021_MOGONET}. Trained on mRNA expression, DNA methylation, and miRNA expression data, MOGONET treats each omics type as a distinct view of the same patient cohort and constructs sample similarity networks to exploit inter-sample relationships. MOGONET outperformed several supervised multi-omics integration baselines for classifying the breast invasive carcinoma subtype and identified omics-specific biomarkers associated with different subtypes.

The above methods show a broader shift in cancer subtyping toward models that move beyond feature-level correlations, whether through causal structure learning or multi-view graph-based integration, to produce classifications more closely tied to underlying biology.

\subsection{Prognosis: predicting what interventions prevent}

The Knowledge-aware Multi-label Network (KAMLN) addresses the limitation of existing postoperative complication prediction models, which most treat complications as independent events, ignoring the fact that they often trigger one another \cite{Yuan_2024_KAMLN}. KAMLN constructs a heterogeneous knowledge graph that encodes potential causal relationships among complications, then designs a neural network architecture based on this graph to learn inter-complication dependencies \cite{Yuan_2024_KAMLN}. Using a cohort of patients with lung cancer surgery, KAMLN outperformed baseline methods in multi-label prediction, with SHAP analysis that identified lymph node dissection as having a significant impact across multiple complication groups \cite{Yuan_2024_KAMLN}. Although further external validation is needed, explicit encoding of prior medical knowledge into the network structure offers a route toward more fine-grained surgical risk stratification than conventional single-label approaches \cite{Yuan_2024_KAMLN}.

For the progression from mild cognitive impairment to dementia, Behnam et al. constructed temporal graph representations of individual patients from longitudinal clinical data, including genotypes, biomarkers, and chronic disease histories, and applied a temporal graph convolutional network to predict MCI transitions \cite{behnam2024causal}. Their causal explanation method, which identifies subgraphs of variables that causally drive predictions, outperformed existing explanation techniques in stability, accuracy, and faithfulness \cite{behnam2024causal}. The resulting causal subgraph highlighted hypertension, arrhythmia, congestive heart failure, coronary artery disease, stroke, lipid-related conditions, and sex as informative variables, with sex differences identified as a strong factor in progression pathways \cite{behnam2024causal}.

\subsection{Treatment selection: mechanism-based therapy}

Drug recommendation in the era of polypharmacy demands methods that can separate genuine therapeutic effects from statistical artefacts of prescribing behaviour. CRec addresses this by partitioning drug molecular graph representations into causal and non-causal components using a learned atom-level mask conditioned on each patient's health state \cite{Zhang_2023_CRec}. A proton pump inhibitor, for example, might appear statistically associated with improved cardiovascular outcomes, not due to any direct cardioprotective mechanism, but because it can be co-prescribed alongside aspirin. CRec applies causal intervention to block such confounded pathways, so that recommendations reflect mechanisms instead of co-prescription patterns. In the MIMIC-III and MIMIC-IV benchmarks, this approach improves the accuracy of the recommendation over methods based on rule-based molecular substructure extraction \cite{Zhang_2023_CRec}.

Where CRec operates at the molecular structure level, CausalMed shifts to the patient side  \cite{li2024causalmed}. Standard recommendation systems learn set-to-set mappings between a patient's full diagnostic profile and their medication list, an approach that obscures which drug treats which condition. CausalMed replaces these co-occurrence associations with point-to-point causal links learnt through score-based causal discovery that resolves confounding pathways between diseases, procedures, and medications. A Dynamic Self-Adaptive Attention mechanism then captures how the same diagnosis affects patients differently depending on their overall health state, producing representations that are personalised rather than population-averaged. On MIMIC-III and MIMIC-IV datasets, CausalMed achieved the highest recommendation accuracy while drug-drug interaction rates were comparable to those of safety-focused baselines, balancing a trade-off that previous methods addressed only in part \cite{li2024causalmed}.

PDGrapher uses protein-protein interaction networks as approximations of causal graphs to predict combinatorial therapeutic targets capable of reversing disease phenotypes \cite{Roohani_2024_PDGrapher}. Rather than learning how perturbations alter cell states and then exhaustively searching for matches, the standard indirect approach, PDGrapher solves the inverse problem: given a diseased gene expression profile and a desired treated state, it predicts which genes a perturbagen should target to achieve that transition. Across nine cell lines with chemical perturbations, PDGrapher identifies effective perturbagens in a greater proportion of test samples than competing methods, and shows competitive performance on ten genetic perturbation datasets \cite{Roohani_2024_PDGrapher}. The predicted target sets identify multiple genes whose combined perturbation is expected to shift a disease state toward a treated profile, with predicted targets falling closer to ground-truth targets in the PPI network than would be expected by chance. These approaches mark a shift from “prescribe what worked for similar patients” to treatment selection grounded in causal mechanisms.

\subsection{Continuous monitoring: real-time causal inference}

Biomedical monitoring produces dense streams of temporal data. In intensive care, hundreds of variables are tracked continuously, creating high-dimensional time series with complex interdependencies. Conventional monitoring systems generate frequent false alarms, which often lack clinical relevance and contribute to alarm fatigue and missed critical events \cite{tsien1997poor, schmid2013patient}.

The CiGNN framework for cuffless blood pressure estimation shows how causal reasoning can improve physiological monitoring \cite{Liu_2024_CiGNN, Liu_2023_Causal}. Rather than using all available features from electrocardiogram and photoplethysmogram signals, CiGNN first applies causal discovery to identify features causally related to blood pressure variation. This reduces a large pool of candidate features to a smaller causal subset, improving interpretability and focusing the model on physiologically relevant inputs. The identified causal features tracked blood pressure changes more effectively, although the physiological mechanisms linking these features with blood pressure remain uncharacterised \cite{Liu_2024_CiGNN}. Independent work on PPG pulse wave analysis suggests that several of the morphological features favoured by such causal methods reflect arterial stiffness and vascular compliance \cite{charlton2022assessing}, but confirming these mechanistic links within a causal discovery framework is an open question.

Missing data is a persistent reality in clinical monitoring, and spurious correlations introduced by unobserved confounders threaten model reliability. The Casper framework addresses this in spatiotemporal time series imputation, using front-door adjustment to block non-causal backdoor paths induced by unknown confounders \cite{Jing_2024_Casper}. It enforces sparse causal relationships among embeddings, with theoretical analysis showing convergence to gradient-based explanations. Evaluated on air quality and traffic sensor networks, Casper achieved improved imputation quality with lower variance compared to baselines, and attained the highest causal discovery accuracy among comparison methods on quasi-realistic gene expression data from the DREAM-3 benchmark \cite{Jing_2024_Casper}. The methodology has not yet been applied to clinical monitoring data, but the underlying problem, missing observations in spatiotemporal sensor networks corrupted by unknown confounders, arises routinely in healthcare settings where sensor dropout, irregular sampling, and environmental noise affect physiological time series.

Cascading events across interconnected physiological systems present a further challenge. Tracing how local perturbations propagate through coupled networks is a structural problem shared across domains, and two recent lines of work address it from complementary angles. In sepsis, organ dysfunction can cascade through the cardiovascular, respiratory, and renal systems in patterns that conventional severity scores fail to capture. Feng et al. modelled the temporal trajectory of inter-organ coupling using a deep temporal graph clustering framework trained on 10,181 ICU patients and externally validated on a further 6,208 \cite{feng2025subphenotyping}. Their model found three phenotypes distinguished by their coupling dynamics: sustained synchronous improvement across organ systems, persistent decoupling, and a rapid transition from early asynchrony to synchronised deterioration. These phenotypes predicted in-hospital mortality more accurately than SOFA or APACHE II scores alone and revealed differential responses to fluid resuscitation strategies, suggesting that the temporal structure of inter-organ interactions carries prognostic information beyond what static or aggregate scoring captures. From the engineering side, the REASON framework uses hierarchical graph neural networks to model both intra-level and inter-level non-linear causal relationships, tracing how faults cascade through interdependent network structures \cite{Wang_2023_Hierarchical}. In both settings, a perturbation at one node can propagate to dependent subsystems at different hierarchical levels, and root cause localisation requires jointly modelling within-network and cross-network causal pathways. Adapting such hierarchical causal discovery methods to physiological monitoring, where organ systems form precisely this kind of interdependent network, remains an open and interesting direction. In sepsis specifically, the phenotype-dependent fluid responses illustrate why this matters. If a real-time system could trace an emerging renal perturbation to its upstream haemodynamic cause rather than treating it as an isolated organ event, management could be tailored to the patient's coupling pattern rather than applied according to a uniform protocol. Whether current methods are computationally fast enough and sufficiently validated to support such decisions at the bedside is as of yet unresolved.

\begin{table}[htbp]
\centering
\small
\caption{Comparative analysis of representative causal graph neural network methods by clinical capability. Methods marked with * were developed and evaluated outside clinical settings but are included for their translational relevance to healthcare applications.}
\label{tab:clinical_methods}
\begin{tabular}{@{}p{0.14\textwidth}p{0.16\textwidth}p{0.28\textwidth}p{0.32\textwidth}@{}}
\toprule
\textbf{Task} & \textbf{Methods} & \textbf{Key Innovation} & \textbf{Potential Clinical Impact} \\
\midrule
Diagnostic subtyping & CI-GNN \cite{Zheng_2024_CIGNN}, BPI-GNN \cite{Zheng_2024_BPI} & Causal subgraph disentanglement via conditional mutual information; prototype learning for disorder subtyping & Psychiatric subtypes with distinct symptom profiles; cross-site stability on multi-site datasets \\
\addlinespace
Cancer classification & CautionGCN \cite{Wang_2024_CautionGCN}, MOGONET \cite{Wang_2021_MOGONET} & Causal gene selection via multi-head autoencoder; GCN-based multi-omics integration with cross-view correlation discovery & Reduced feature sets via causal gene selection; multi-omics integration across RNAseq, DNA methylation, and copy number variation \\
\addlinespace
Seizure prediction & DIGLN \cite{lian2025dynamic} & Patient-specific dynamic graph structure learning from intracranial EEG & Joint network--temporal modelling on Freiburg iEEG; external validation on larger cohorts still needed \\
\addlinespace
Complication prediction & KAMLN \cite{Yuan_2024_KAMLN} & Heterogeneous knowledge graph encoding potential inter-complication causal dependencies & Multi-label prediction on lung cancer cohort \\
\addlinespace
Drug recommendation & CRec \cite{Zhang_2023_CRec}, CausalMed \cite{li2024causalmed} & CRec: causal intervention separating causal from spurious molecular components via learnt atom-level masks, conditioned on patient health state. CausalMed: point-to-point causal links with score-based causal discovery replacing co-occurrence mappings; dynamic self-adaptive attention mechanism & Personalised recommendations grounded in learnt causal disease-medication relationships on MIMIC-III/IV; best accuracy with acceptable DDI rates \\
\addlinespace
Combination therapy* & PDGrapher \cite{Roohani_2024_PDGrapher} & Inverse causal inference on protein--protein networks & Direct perturbagen prediction for phenotype reversal across cancer cell lines \\
\addlinespace
Blood pressure monitoring & CiGNN \cite{Liu_2024_CiGNN} & FCI- and CGNN-based causal graph inference from wearable sensors, used as GNN topology & Causal PPG features beyond pulse transit time; robust under acute BP perturbations \\
\addlinespace
Spatiotemporal imputation* & Casper \cite{Jing_2024_Casper} & Frontdoor adjustment blocking unknown spatiotemporal confounders & Improved imputation quality with lower variance on air quality and traffic sensor networks; highest causal discovery accuracy on gene expression data. Methodology applicable to clinical monitoring with sensor dropout and irregular sampling \\
\addlinespace
Cognitive decline & CXGNN \cite{Behnam_2025_ECCV, behnam2024causal} & Structural causal models over graph structures with neural causal model parameterisation for causal GNN explanation & Causal subgraph identifying vascular risk factors and sex as drivers of MCI-to-dementia progression; consistent with epidemiological evidence \\
\bottomrule
\end{tabular}
\begin{tablenotes}
\small
\item Abbreviations: CI-GNN, Causality-Inspired Graph Neural Network; BPI-GNN, Brain Prototype-Inspired GNN; CautionGCN, Causal Attention Graph Convolutional Network; MOGONET, Multi-Omics Graph cOnvolutional NETwork; DIGLN, Dynamic Instance-level Graph Learning Network; KAMLN, Knowledge-Aware Multi-label Network; CRec, Conditional Causal Representation Learner; CausalMed, Causal Medicine recommendation framework; PDGrapher, Perturbation-Driven Grapher; CiGNN, Causality-informed Graph Neural Network; Casper, Causality-Aware Spatiotemporal Imputation; CXGNN, Causal Explainer GNN; FCI, Fast Causal Inference; CGNN, Causal Generative Neural Network; MCI, Mild Cognitive Impairment; PPG, Photoplethysmogram; DDI, Drug-Drug Interaction; BP, Blood Pressure.
\end{tablenotes}
\end{table}

\section{Towards causal digital twins}

Table~\ref{tab:synthesis} summarises how the methods described in Section 3 and validated across domains in Section 4 converge toward patient-specific causal inference. Digital twins originated in manufacturing as virtual representations of physical systems \cite{grieves2023digital} and have since gained traction in healthcare, where they offer an approach for building dynamic, patient-specific computational models \cite{EmmertStreib_2025_DigitalTwins}. Unlike static predictive models, digital twins remain connected to their physical counterparts, updating their representations as new patient data become available at clinical time points. Emmert-Streib and colleagues characterise four core capabilities of digital twins: explainability, grounded in mechanistic model structures that are interpretable in biological terms; intervenability, allowing simulation of therapeutic interventions before they are administered; learnability, whereby the model refines itself as patient trajectories develop over time; and diversability, which uses parameter uncertainties to generate ensemble predictions that quantify prognostic confidence \cite{EmmertStreib_2025_DigitalTwins}. Several perspectives have outlined how these ideas apply across cardiovascular medicine \cite{Coorey_2021_Cardiovascular}, immunology \cite{Laubenbacher_2022_Roadmap}, and oncology \cite{Wu_2022_Oncology_DT}, while closed-loop systems such as the artificial pancreas \cite{Kovatchev_2018_Artificial} demonstrate that real-time computational models maintaining dynamic patient feedback can achieve clinical deployment, offering a partial precedent for narrowly scoped digital twins.

\begin{table}[htbp]
\centering
\small
\caption{Mapping methodological capabilities to clinical applications and digital twin components. Clinical applications reflect demonstrated uses where available; where methods were developed outside clinical settings, we indicate translational potential discussed in the text.}
\label{tab:synthesis}
\begin{tabular}{@{}p{0.18\textwidth}p{0.26\textwidth}p{0.22\textwidth}p{0.24\textwidth}@{}}
\toprule
\textbf{Capability (\S3)} & \textbf{Clinical Application (\S4)} & \textbf{Digital Twin Role (\S5)} & \textbf{Methods} \\
\midrule
Causal disentanglement & Psychiatric classification (CI-GNN); psychiatric subtyping (BPI-GNN); drug recommendation on ICU data (CRec); potential for cancer subtyping and other diagnostic contexts & Isolating patient-specific mechanisms & DisC \cite{fan2022debiasing}, CI-GNN \cite{Zheng_2024_CIGNN}, BPI-GNN \cite{Zheng_2024_BPI}, CRec \cite{Zhang_2023_CRec} \\
\addlinespace
Intervention prediction & Perturbagen target prediction on cancer cell lines (PDGrapher); potential for treatment selection and therapeutic testing & Virtual therapeutic testing & Ze\v{c}evi\'{c} et al.\ \cite{Zecevic_2021_Relating}, RC-Explainer \cite{Wang_2022_RCExplainer}, CAL \cite{Sui2022causal}, PDGrapher \cite{Roohani_2024_PDGrapher}, CXGNN \cite{Behnam_2025_ECCV, behnam2024causal} \\
\addlinespace
Counterfactual generation & Cognitive decline prognosis; epilepsy prediction & Patient-specific outcome simulation & CLEAR \cite{Ma_2022_Clear}, Huang et al.\ \cite{huang2025local} \\
\addlinespace
Invariant learning & Potential for cross-institutional deployment & Generalisation across clinical contexts & DIR \cite{wu2022discovering}, IGCL-GNN \cite{Liu_2024_IGCLGNN} \\
\addlinespace
Causal fairness & ICU outcome disparity analysis; treatment allocation equity in coronary artery disease & Equity-preserving simulation & Plecko et al.\ \cite{plecko2025algorithmic}, Zhang et al.\ \cite{zhang2024causal}, path-specific methods \cite{Plecko_2024_Causal_Fairness} \\
\bottomrule
\end{tabular}
\end{table}

Most current digital twin implementations, however, are based on physics-based organ simulations or data-driven models that learn patient-specific parameters, and neither framework successfully includes formal causal reasoning as would be needed for applications in health \cite{de2025challenges}. Physics-based approaches offer mechanistic interpretability but become unwieldy for multisystemic diseases that span molecular, cellular, and organ-level interactions \cite{de2025challenges, sadee2025medical}. Data-driven approaches handle this complexity better, but because they learn from observational associations, they cannot reliably predict outcomes under novel interventions or in patient populations that differ from those in the training data \cite{Pearl_2009_Causality}. Causal graph neural networks offer one route toward bridging this gap. By embedding structural causal model assumptions into neural architectures, they can in principle support interventional queries through the do-calculus \cite{Pearl_2009_Causality} and counterfactual reasoning at the individual patient level, while invariance-based learning objectives \cite{peters2016causal, Arjovsky_2019_Invariant} can improve robustness across heterogeneous clinical settings. Together, these ideas point toward what we term Causal Digital Twins. However, realising this depends on assumptions about the causal structure that remain difficult to verify from clinical data alone.

\subsection{Patient-specific mechanistic simulation}

Causal inference and graph learning could together support a longer-term vision for precision medicine. Patient-specific Causal Digital Twins would represent dynamic computational models in which clinicians perform in silico experiments to test interventions before clinical administration. Such systems would need to integrate a patient’s multi-omics profile encompassing genomics, transcriptomics, proteomics, and metabolomics; longitudinal imaging capturing organ structure and function over time; clinical history, including treatments, responses, and adverse events; and knowledge graphs encoding established biological mechanisms \cite{Regev_2017_Human, Topol_2019_High, Nicholson_2023_Constructing, chandak2023building}. A causal graph neural network architecture could learn patient-specific parameterisations of general biological mechanisms by combining observational patient data with prior mechanistic knowledge, and simulate therapeutic interventions through the do-operator framework \cite{Pearl_2009_Causality}. By setting treatment variables, propagating effects through learnt causal pathways, and predicting outcomes across biological scales, such a system would generate predictions grounded in assumed causal structure.

Consider molecularly targeted oncology, where treatment decisions already depend on genomic profiling, but where the number of actionable alterations and candidate therapies continues to grow \cite{sicklick2019molecular}. A Causal Digital Twin could encode the patient’s molecular profile, mutational, transcriptomic, and proteomic, within a causal graph that captures known pathway dependencies. The clinician could then simulate how candidate strategies would propagate through the patient’s specific network by comparing, for example, the expected durability of monotherapy with combination or sequential regimens, estimating likely resistance trajectories, and weighing predicted efficacy against anticipated side effects (Figure \ref{fig:causal_failure_solutions}). These predictions would remain conditional on the assumed causal structure, and their clinical value would depend on prospective validation. The aim is not to replace clinical judgement but to help narrow the therapeutic search space when multiple options exist, a situation that is becoming routine as genomic profiling identifies increasing numbers of co-occurring targetable alterations.

Current methodological work provides some support for this vision, though none yet delivers a complete causal digital twin. At the theoretical level, graph neural network message passing can be formally related to do-calculus computations, thereby establishing that graph architectures are, in principle, capable of interventional reasoning \cite{Zecevic_2021_Relating}. Causal deep learning can span structural, parametric, and temporal dimensions, with applications to treatment effect estimation and counterfactual outcome prediction in healthcare \cite{berrevoets2023causal}. On the applied side, counterfactual explanation methods such as CLEAR generate alternative graph-structured scenarios using variational autoencoder mechanisms that promote causal consistency \cite{Ma_2022_Clear}, and PDGrapher uses protein-protein interaction networks as causal graph approximations to predict the effects of therapeutic perturbations \cite{Roohani_2024_PDGrapher}. Multimodal integration frameworks such as MOGONET and MoCaGCN demonstrate how diverse omics data can be integrated using graph-based architectures \cite{Wang_2021_MOGONET, Li_2024_MoCaGCN}. Separately, causal fairness frameworks provide tools for tracing discrimination along specific causal pathways, which would be important for any clinically deployed system \cite{Plecko_2024_Causal_Fairness}. Bringing these components together into a system that jointly performs causal discovery, multi-modal integration, interventional prediction, and fairness-aware reasoning for individual patients remains an open challenge, scientifically and in engineering, but the theoretical and methodological foundations are coming into place.

\subsection{The LLM-causal graph neural network synergy}

Large Language Models (LLMs) have been shown to be robust in learning from unstructured clinical text such as progress notes, radiology reports, pathology narratives, and can extract entities, relationships, and temporal sequences \cite{thirunavukarasu2023large, mesinovic2025explainability}. However, they remain limited in causal reasoning. Benchmark testing of formal causal inference has found that LLMs perform poorly on causal queries, often relying on surface-level correlations from training data rather than on principled causal logic \cite{jin2023cladder}. These models can identify that two variables co-occur, but they cannot reliably determine whether one causes the other.
Causal graph neural networks can help make up for this missing capability. The trade-off is that they might depend on well-specified graph topologies and fare poorly with the kind of unstructured, free-text information that LLMs handle with ease \cite{mesinovic2025explainability}. A combined architecture could therefore use LLMs to structure raw clinical text into graph-amenable representations, and causal graph neural networks could then use the resulting structures to discover mechanistic relationships.

Novel adverse drug reactions may surface in clinical narratives well before structured reporting systems accumulate sufficient signal for detection. Recent work has shown that causal knowledge graphs integrating disease progression pathways, drug indications, and side-effect databases can identify both known and previously undocumented adverse drug reactions from large-scale observational cohorts \cite{toonsi2026causal}. An LLM could parse these records to identify temporal coincidences between drug exposures and emerging symptoms and generate candidate causal hypotheses. A causal graph neural network could then evaluate such hypotheses against knowledge graphs encoding drug-target and protein-protein interactions, applying identifiability criteria and sensitivity analyses from the causal inference framework \cite{Pearl_2009_Causality, Hernan_2020_Causal} to determine whether the proposed relationship is likely to be confounded. The individual components, LLM-based clinical text mining and causal knowledge graphs for drug safety, are active areas of development, and could be achievable in the near-term.

The aforementioned hybrid architectures must contend with well-documented LLM limitations. Evaluations of reference generation have found concerning hallucination rates across models, with even non-fabricated outputs frequently failing to meet specified inclusion criteria \cite{Chelli_2024_Hallucination}. More broadly, benchmarking across biomedical NLP tasks shows that while LLMs perform well in reasoning-intensive question answering, they lag behind fine-tuned domain-specific models in information extraction in few-shot settings, and qualitative analysis reveals persistent gaps and inconsistencies in their outputs \cite{Chen_2025_BioNLP}. Early work on knowledge graph-augmented generation in biomedicine shows that grounding LLM outputs against structured knowledge graphs could potentially reduce hallucination rates in text-based tasks \cite{lavrinovics2025knowledge}. Causal graph neural networks operating over similar graph structures could extend this grounding from factual consistency to mechanistic validity.

\subsection{Implementation requirements}

Making the Causal Digital Twin vision work in practice depends on progress across computation, data integration, validation, and regulation. On the computational side, causal graph neural network methods could incur considerable overhead due to causal discovery and multi-environment optimisation. Hierarchical causal discovery that takes advantage of biological modularity could reduce this burden, as could amortised inference schemes that pay the cost of training once, but produce fast predictions at deployment. Cloud infrastructure with distributed graph processing would help meet the necessary scale while keeping response times more clinically acceptable. Integrating multi-modal biomedical data is a separate and substantial challenge. Clinical AI systems have to handle genomic, transcriptomic, proteomic, metabolomic, imaging, free-text, structured EHR, and physiological monitoring data, each with its own preprocessing, quality control, and normalisation requirements \cite{acosta2022multimodal, Topol_2019_High}. Temporal alignment is also difficult, as genome sequences are measured only once per patient, whereas physiological signals may be sampled thousands of times per second. All of these data streams need to map onto a shared causal structure. Federated learning, however, is a way to train models across institutions without centralising sensitive patient data, addressing both privacy regulation and the practical reality that most health data cannot leave the hospital that collected it \cite{rieke2020future}. Validation of digital twin predictions needs clear evidentiary standards. A multimodal triangulation framework would offer a starting point, combining assessments of biological plausibility, replication across patient cohorts, natural experiments where they exist, and prospective comparisons of predictions against observed outcomes.

On the regulatory side, frameworks for in silico experimentation are still catching up. The FDA has published guidelines on evaluating the credibility of modelling in medical device (including AI) submissions  \cite{FDA_2023_CMS, FDA_AI}, and both the FDA and the EMA could accept in silico evidence for device evaluation \cite{viceconti2021silico}. However, extending these frameworks to patient-specific treatment optimisation requires new guidance to address the unique challenges of personalised mechanistic modelling. Questions of liability when predictions are wrong, intellectual property over patient-specific models, and equitable access to computationally intensive tools all need policy attention. None of this will happen without collaboration between developers of causal inference methods, clinicians who understand what the tools need to do, biomedical researchers who can provide mechanistic knowledge, and regulators willing to develop appropriate oversight. The technical building blocks exist. Whether they come together into something clinically useful depends considerably on organisational and policy progress.

\section{Challenges and path forward}

\subsection{Computational reality}

Biological systems help with costs as they are modular, with interactions mostly confined to local functional units \cite{hartwell1999molecular}. Causal discovery algorithms can exploit this sparsity through hierarchical decomposition, resolving coarse relationships between subsystems before refining edges within each module. Amortised inference can also help, where a neural model trained on simulated causal datasets (in this case, semisynthetic gene expression data) learns to predict the graph structure from new observations in a single forward pass, bypassing combinatorial search entirely \cite{lorch2022amortized}. None of this implies that every clinical prediction task needs causal machinery. Where distribution shift is minimal, and intervention modelling is unnecessary, standard associational models are cheaper and often sufficient.

\begin{tcolorbox}[colback=blue!5!white,colframe=blue!75!black,title=\textbf{Box 2: Computational complexity and scaling strategies}]
\textbf{The scaling challenge.}
The number of possible directed acyclic graphs grows super-exponentially with the number of variables, making exhaustive causal discovery intractable beyond modest network sizes~\cite{Pearl_2009_Causality}. Constraint-based algorithms can reduce this cost under sparsity assumptions, but still scale poorly with set size. For context, you could represent human cells as the expression of around 20,000 protein-coding genes~\cite{Regev_2017_Human}, keeping in mind that clinical deployment typically concerns second-level inference.

\vspace{0.2cm}
\begin{center}
\begin{tabular}{lll}
\toprule
\textbf{Strategy} & \textbf{Key idea} & \textbf{Example} \\
\midrule
Structural priors & Use known interaction networks as & PDGrapher~\cite{Roohani_2024_PDGrapher} \\
 & causal proxies, bypassing discovery & \\
Amortised inference & Train once on simulated data, then & AVICI~\cite{lorch2022amortized} \\
 & predict structure in a forward pass & \\
\bottomrule
\end{tabular}
\end{center}
\vspace{0.2cm}

\textbf{Empirical reference point.}
CI-GNN performs causal subgraph identification on 116-node brain networks in 32.8ms per instance at inference~\cite{Zheng_2024_CIGNN}, suggesting that GNN-based causal reasoning could meet latency requirements on moderately sized graphs. The challenge is reaching larger scale networks at comparable speed.
\end{tcolorbox}

\subsection{Circumstances of failure}

Recent large-scale empirical work challenges optimistic assumptions about causal generalisation. In an evaluation of 16 prediction tasks in the domains of healthcare, employment, education, and social policy, each with multiple deployment environments for out-of-distribution testing \cite{Nastl_2024_NeurIPS}, predictors using all available features, regardless of causality, achieved superior accuracy in-domain and out-of-domain compared to predictors restricted to causal features. Even the absolute accuracy drop from training to deployment domains was not better for causal predictors than for models exploiting all correlational patterns. Causal machine learning methods specifically designed for domain generalisation did not improve upon standard predictors trained on arguably causal features, with their performance typically falling between models trained on conservative and inclusive causal feature selections. Causal discovery algorithms either failed to execute on real-world datasets or selected variables, offering no advantage over expert-curated selections. Complementary synthetic experiments confirmed that a substantial distributional shift is required for causal features to achieve superior out-of-domain accuracy, suggesting that when shifts are modest, the predictive information lost by discarding non-causal features may outweigh the robustness gained.

Sample size constraints pose challenges for causal graph neural networks in neuroimaging applications. Brain connectivity studies can involve high-dimensional graphs comprising hundreds to thousands of nodes, with relatively small patient cohorts, thereby posing a substantial risk of overfitting \cite{Mohammadi_2025_BrainSci}. Neural network models can exhibit greater variance than conventional machine learning methods on such limited data, and regularisation strategies developed for standard GNNs do not necessarily transfer to causal architectures with additional structural constraints. The BrainGB benchmark also showed that sophisticated GNN architectures such as BrainGNN could exhaust memory on large-scale datasets, and although GNNs outperformed shallow baselines even on some small datasets, performance gains were inconsistent and dataset-dependent \cite{Cui_2023_BrainGB}. The gap between the complexity of the model that causal reasoning requires and the availability of data common in clinical neuroimaging remains an obstacle.

Unobserved confounding remains a limitation that architectural innovation struggles to overcome (Figure~\ref{fig:causal_hierarchy}B). Observational studies in biomedicine are commonly affected by confounding and selection bias, and the development of health intervention models from such data is problematic regardless of method sophistication, not least because electronic medical records do not explicitly record contextual knowledge such as why one drug was prescribed over another \cite{Prosperi_2020_Nmi}. Automated causal structure learning itself requires the assumption of `causal sufficiency' that there are no unmeasured common causes and no selection bias, but healthcare data often violate this condition. When treatment decisions depend on unmeasured clinical judgement, patient preferences, or institutional practices, even sophisticated causal graph neural network architectures risk learning spurious relationships that appear causally valid under their structural assumptions but are driven by residual confounding. The CausalBench benchmark evaluated network inference methods on large-scale single-cell perturbation data using biologically-motivated and distribution-based interventional metrics and found that poor scalability limits the performance of current methods \cite{Chevalley_2025_CausalBench}. In an initial evaluation, established interventional methods did not outperform observational-only approaches. Methods subsequently developed through a community challenge successfully leveraged interventional data and achieved substantially higher performance, but this required designs that avoided common assumptions such as acyclicity and causal sufficiency, suggesting that standard causal discovery frameworks may need substantial adaptation for gene regulatory network inference from single-cell data.

The above empirical limitations show that causal graph neural networks should be deployed with appropriate epistemic humility. Methods are most justified when specific failure modes, distribution shift between training and deployment, confounded treatment effects requiring debiasing, or counterfactual personalisation demanding patient-specific simulation, are clinically relevant and sufficiently severe to justify the additional computational overhead. When these conditions are absent, correlation-based approaches remain appropriate and more practical.

\subsection{The validation crisis}

Establishing trustworthy causal conclusions from observational healthcare data is perhaps the most difficult challenge. Standard machine learning evaluations, including cross-validation and held-out test sets, as well as those used in treatment effect modelling, effectively assess predictive performance but provide insufficient evidence to validate causal claims \cite{feuerriegel2024causal, alaa2019validating, Pearl_2009_Causality}. Standard machine learning evaluation metrics, including accuracy, area under the curve, and calibration measure association rather than causation. A model can achieve excellent predictive performance while encoding spurious correlations that would fail under distribution shift \cite{subbaswamy2020development, Peters_2016_Causal_Inference}. Ground truth causal structures remain unknown for most biomedical systems, thus direct validation of learnt causal graphs is extremely difficult.

Table~\ref{tab:assumptions} sets out the assumptions underlying each methodological class reviewed in this Perspective. Violation of these assumptions, especially causal sufficiency in the presence of unmeasured confounding, invalidates the causal interpretation of model outputs regardless of predictive performance. The tiered evidentiary framework provides a mechanism for calibrating confidence: Tier 1 methods make no causal claims and do not require these assumptions; Tier 2 methods require that assumptions be plausible, supported by domain knowledge; Tier 3 claims demand empirical validation that the assumptions hold in the specific application context.

The validation crisis (Figure~\ref{fig:causal_hierarchy}A) shows that the methods clustered in the high-correlation, low-intervention quadrant achieve impressive retrospective metrics but might fail at prospective deployment. This ``causal gap" quantifies the expected performance loss when correlation-based models are applied to interventional queries. This underscores the need for validation frameworks that extend beyond cross-validation to include natural experiments, propensity-matched analyses, and prospective intervention studies \cite{feuerriegel2024causal,Hernan_2016_Using_Big_Data}.

Multi-modal evidence triangulation can serve as a framework for establishing causal validity through convergent evidence across complementary modalities \cite{Lawlor_2016_Triangulation, feuerriegel2024causal}. Biological plausibility assessment would evaluate whether identified causal relationships align with established pathophysiological mechanisms and biochemical pathways and provide mechanistic coherence testing against biomedical knowledge \cite{Hill_1965_Environment}. The MoCaGCN framework integrates prior knowledge to constrain the learnt causal gene graphs (using a Bayesian approach) and ensure that the identified relationships respect established gene-gene interactions \cite{Li_2024_MoCaGCN}. Replication among independent cohorts helps test whether causal relationships remain stable across diverse populations, healthcare systems, and temporal periods. The CiGNN blood pressure monitoring framework, for example, was evaluated on several datasets and demonstrated consistent performance in both healthy and patient populations \cite{Liu_2024_CiGNN}.

The validation design also has to address the resolution in which causal claims are made. In the interventionist framework, causal relationships are defined relative to interventions that could, in principle, be performed. When computational nodes aggregate multiple biological entities, such as the “prefrontal cortex,” which comprises dozens of subregions, or “inflammation,” which encompasses multiple cytokine cascades, experimental validation becomes ambiguous. Interventions targeting specific sub-components may not test the aggregate causal claim, whereas aggregate-level interventions may be infeasible or imprecise. This is a resolution mismatch between causal graphs operating at the pathway level, which require experimental designs that modulate entire pathways coherently, and gene-level graphs, which permit targeted perturbation experiments. Explicit specification of node semantics in causal graph reporting, analogous to intervention definitions in clinical trial protocols, would help clarify what experiments could validate the claims and at what biological scale the causal relationships hold.

External validation designs can further strengthen causal claims by providing estimates for benchmarking against observational models. Policy discontinuities, instrumental variables, and regression discontinuity designs allow the estimation of causal effects with minimal confounding and serve as benchmarks against which observational causal models can be validated. Comparison against randomised trial data, when accessible, offers the most direct form of validation by testing whether estimated treatment effects align with experimentally derived outcomes  \cite{feuerriegel2024causal, Hernan_2016_Using_Big_Data}. Sensitivity analyses can quantify robustness to violations of causal assumptions, particularly unmeasured confounding, through E-value calculations \cite{feuerriegel2024causal, VanderWeele_2017_Sensitivity} and omitted variable bias frameworks that assess how strong unobserved confounders would need to be to alter conclusions \cite{Cinelli_2020_Making_Sense}.

We propose tiered evidentiary standards that distinguish the strength of causal evidence. Tier 1 encompasses causally-inspired methods that employ causal concepts in architectural design but make no explicit causal claims about learnt relationships, such as invariance-based regularisation promoting features stable across environments. Tier 2 includes causally structured methods that learn causal structures or estimate causal effects, validated by consistency with external knowledge rather than by novel discovery, exemplified by cancer subtyping models that recover known oncogenic pathways. Tier 3 represents causally validated methods that claim the discovery of previously unknown causal relationships, requiring triangulation of biological plausibility with the proposed mechanism, replication across two or more independent cohorts, validation through experiments or prospective intervention studies, and sensitivity analysis demonstrating robustness using E-values for strong claims.

\subsection{From methods to clinical practice}

The gap between methodological sophistication and clinical interpretability remains a barrier to deployment. Clinicians often connect trust in predictions with explanations that they can understand, verify against their own expertise, and combine with clinical reasoning \cite{Murdoch_2021_Definitions, tonekaboni2019clinicians}. For GNNs specifically, it can be difficult to ensure that the patterns discovered are both interpretable and clinically relevant \cite{Mohammadi_2025_BrainSci, paul2024systematic}. The representations these networks learn can be difficult to map onto clinical concepts directly, and bridging that disconnect requires dedicated translation frameworks. One way to close this gap is to build domain knowledge directly into model design. Rather than learning causal structures purely from data, hybrid approaches can incorporate established biological mechanisms as structural priors, constraining the architecture around known pathway relationships. In cancer subtyping, for example, MoCaGCN can combine causal structure learning with gene-gene interaction priors to guide graph construction, so that the model's learnt representations remain grounded in biology while still allowing data-driven discovery \cite{Li_2024_MoCaGCN}. Recent work on causal subgraph disentanglement has shown that graphs can be decomposed into causal and non-causal components and improve both generalisation and interpretability \cite{fan2022debiasing}. CI-GNN uses a Granger causality-inspired mechanism to isolate sparse subgraphs of functional connectivity in brain networks that are causally related to psychiatric diagnoses, with identified connections reported to align with clinical evidence from neuroimaging studies \cite{Zheng_2024_CIGNN}.

Clinician-in-the-loop validation offers a check on the outputs, with domain experts assessing whether identified causal mechanisms are biologically accurate and clinically coherent. This kind of iterative refinement, where data-driven discovery is tested against expert knowledge, is important for building clinical trust, although formalising it within deployment pipelines remains an open problem. On the regulatory side, existing frameworks have not kept pace. The FDA has issued guidance on real-world evidence for medical devices \cite{FDA_2021_AI, FDA_AI} and has moved towards a regulatory pathway for AI/ML-based software as a medical device, but it focuses on predictive performance and safety monitoring rather than validating causal or counterfactual claims derived from observational data. Appropriate oversight for such claims will need to balance innovation with patient safety, establishing evidentiary standards that are rigorous enough to be credible without imposing validation burdens that are practically infeasible.

There is also a risk of ``causal-washing", where methods adopt causal terminology without meeting the evidentiary standards that causal claims actually require. The appeal is easy to understand, causal framing offers stronger theoretical properties and aligns naturally with how clinicians reason, but the gap between causal aspiration and causal validation is wide. In practice, this takes several forms. Some methods use vocabulary such as ``causal representation", ``causal discovery" or ``causal reasoning" while relying on associational quantification that never estimates interventional distributions or tests identifiability conditions \cite{Pearl_2009_Causality, Peters_2016_Causal_Inference}. Others ground their causal claims entirely in predictive performance, without triangulating against biological plausibility, replication, real-world experiments, or sensitivity analyses. Omission of the assumptions on which causal methods depend, faithfulness, causal sufficiency, and positivity, makes it difficult to assess whether the claimed inferences are valid \cite{Schoelkopf_2021_Toward, Peters_2016_Causal_Inference}.

Standardised reporting could help. Similarly, as CONSORT hoped to improve transparency in randomised trials and STROBE did the same for observational studies \cite{Vandenbroucke_2016_Strengthening}, analogous guidelines for causal claims would give reviewers and readers a shared basis for evaluation. Authors should be expected to state their causal assumptions with directed acyclic graphs where appropriate and to specify which causal quantities they are estimating, whether average treatment effects, conditional average treatment effects, or something else. Sensitivity analyses that test robustness to assumption violations should be reported as standard, and findings framed as causal need to be clearly separated from those that are associational. None of this will happen through methodology alone. Funding agencies would need to value validation work over the next novel method and peer reviewers would need to push back on non-validated causal claims.

\begin{tcolorbox}[colback=blue!5!white,colframe=blue!75!black,title=\textbf{Box 3: Illustrative example of multi-modal validation framework}]

\textbf{The amyloid-$\beta$ hypothesis in Alzheimer's disease} \\
Consider a causal graph neural network identifying a directed edge from amyloid-$\beta$ deposition to cognitive decline in Alzheimer's disease, where observational and interventional evidence might point in different directions.\\

\textbf{Evidence triangulation:}
\begin{enumerate}
\item \textbf{Biological plausibility ($S_{\text{bio}} = +1$):} Strong mechanistic support through the amyloid cascade hypothesis and decades of foundational neuroscience research.
\item \textbf{Replication ($S_{\text{rep}} = +1$):} Association replicates consistently across major cohorts, including ADNI, AIBL, and European EADC datasets with stable effect sizes.
\item \textbf{Experiments ($S_{\text{quasi}} = -1$):} Suppose Mendelian randomisation studies using genetic variants affecting amyloid production demonstrated no causal effect on cognitive decline.
\item \textbf{Prospective validation ($S_{\text{prosp}} = -1$):} Assume multiple Phase III clinical trials of amyloid-reducing therapies show minimal cognitive benefit despite substantial plaque reduction.
\item \textbf{Sensitivity analysis ($S_{\text{sens}} = 0$):} Moderate robustness to unmeasured confounding (E-value = 1.8) but substantial uncertainty regarding alternative causal structures.
\end{enumerate}

\textbf{Aggregate evidence score:}
\begin{equation}
\text{AES} = 0.20(+1) + 0.25(+1) + 0.30(-1) + 0.15(-1) + 0.10(0) = 0.00
\end{equation}

\textbf{Interpretation:} The null aggregate score reflects how strong observational associations and biological plausibility can be contradicted by experiments and prospective trials \cite{Lawlor_2016_Triangulation, Hemkens_2018_Interpretation}. The amyloid deposition could be acting as a biomarker of disease progression rather than a causal driver, or the causal dynamics might be too complex for the learned graph to capture, such as threshold effects or developmental windows. \\

\textbf{Clinical implication:} Predictions relying on this causal edge should be treated with caution, amyloid reduction, in this illustrative example, alone may not constitute an effective therapeutic strategy. More broadly, systematic evidence triangulation matters. Compelling observational evidence is not sufficient to justify causal claims when interventional data disagree.

\end{tcolorbox}

\begin{tcolorbox}[colback=blue!5!white,colframe=blue!75!black,title=\textbf{Box 4: Imagined clinical workflow for causal drug recommendation}]

\textbf{Clinical scenario} \\
A 68-year-old patient presents with type 2 diabetes, hypertension, and early-stage chronic kidney disease. The electronic health record shows current medications including metformin and lisinopril. The clinical question: which additional antihypertensive agent would optimise cardiovascular outcomes while protecting renal function?\\

\textbf{Step 1: Graph construction}
\begin{enumerate}
\item Patient node features: demographics, comorbidities, laboratory values (HbA1c, eGFR, potassium)
\item Drug nodes: candidate medications with molecular fingerprints encoding mechanism of action
\item Edges: drug-drug interactions, drug-disease relationships, patient-drug history
\end{enumerate}

\textbf{Step 2: Causal disentanglement (CRec framework, for example)} \\
The model separates drug representations into:
\begin{enumerate}
\item \textit{Causal therapeutic component}: molecular mechanisms affecting blood pressure and renal protection (ACE inhibition, calcium channel blockade, mineralocorticoid antagonism)
\item \textit{Spurious prescription component}: historical prescribing patterns reflecting physician habits, formulary constraints, marketing effects
\end{enumerate}

\textbf{Step 3: Interventional prediction} \\
For each candidate drug $d$, the model computes $P(\text{outcome} | \text{do}(d), \text{patient features})$ rather than $P(\text{outcome} | d, \text{patient features})$, removing confounding from indication bias (sicker patients receiving more aggressive treatment).\\

\textbf{Step 4: Counterfactual comparison}
\begin{center}
\begin{tabular}{lcc}
\toprule
\textbf{Candidate} & \textbf{CV Risk Reduction} & \textbf{eGFR Preservation} \\
\midrule
Amlodipine & 12\% & Neutral \\
Chlorthalidone & 18\% & $-$8\% annual decline \\
Spironolactone & 22\% & $+$4\% preservation \\
\bottomrule
\end{tabular}
\end{center}

\textbf{Step 5: Recommendation with explanation} \\
The system recommends spironolactone with a mechanistic rationale. Mineralocorticoid receptor antagonism reduces cardiac fibrosis (cardiovascular benefit) while counteracting aldosterone-mediated nephron loss (renal protection). This recommendation differs from historical prescribing patterns in which chlorthalidone was more commonly selected, despite evidence of inferior renal outcomes in diabetic nephropathy.\\

\textbf{Clinical validation pathway} \\
Trial evidence supports the underlying causal mechanism. The FIDELIO-DKD trial showed the cardiorenal benefits of mineralocorticoid receptor antagonism in diabetic kidney disease, albeit with finerenone rather than spironolactone \cite{Bakris_2020_FIDELIO}. This provides class-level mechanistic support for the model's recommendation, though drug-specific validation would require further evidence. The difference from standard recommendation systems is that the prediction is grounded in estimated causal drug effects and not just co-occurrence patterns in prescribing histories.

\end{tcolorbox}

We note that healthcare presents additional methodological challenges where causal reasoning is needed, but which exceed the scope of this Perspective. Deep learning models are data-hungry, and neuroimaging cohorts are typically small, and the tension between high-dimensional inputs and limited sample sizes makes overfitting a persistent concern when applied in large scale data such as brain imaging studies \cite{Smucny_2022_DLNeuroimaging}. Multi-centre studies face inter-site heterogeneity introduced by different scanners, protocols, and patient populations which requires dedicated harmonisation strategies \cite{Guan_2021_MultiSite}, and the invariant learning approaches discussed in Section 3.4 do not fully resolve these domain adaptation challenges. Longitudinal treatment regimes introduce time-varying confounding that standard cross-sectional causal graphs cannot accommodate \cite{Keogh_2023_TimeVarying}. Finally, missing data in electronic health records remains widespread yet is poorly handled in existing causal graph neural network models. Recent systematic evidence confirms that no single imputation strategy dominates across missingness scenarios \cite{Ren_2024_MissingEHR}, and integrating principled missing-data methods into graph-based causal frameworks is largely unexplored. Each of these directions calls for continued methodological development.

\section{Conclusion}

Causal graph neural networks offer a route toward AI in healthcare systems that learn biological mechanisms rather than statistical shortcuts. The failures documented throughout this Perspective, performance collapse under distribution shift, perpetuation of discriminatory patterns, and inability to support interventional reasoning, stem from a shared source: models trained on observational associations cannot reliably generalise to the causal questions that clinical decisions often require. The methods reviewed here, from causal disentanglement and invariant learning to interventional prediction and counterfactual generation, begin to address this gap by embedding structural causal assumptions into graph architectures suited to the networked structure of biomedical data.

The barriers to clinical translation remain. Computational costs restrict causal discovery and multienvironment optimisation to offline research settings, far from the real-time demands of clinical deployment. Validating causal claims from observational data is even harder, since standard predictive metrics do not distinguish causal structure from well-fitted spurious correlations, and ground-truth causal graphs are unavailable for most biological systems. The risk of a kind of causal-washing, where methods adopt causal language without the evidentiary rigour that causal claims demand, could erode the field's credibility. The tiered framework we propose, which distinguishes causally-inspired architectures from causally-structured and causally-validated approaches, is one step toward an honest calibration of what different methods actually deliver.

Whether causal graph neural network methods are clinically applicable depends on the progress that no single research group can achieve alone. Scalable causal discovery algorithms that exploit biological modularity, validation standards adapted to observational healthcare data, regulatory frameworks for systems making interventional claims not just purely associational, and clinical trials measuring patient outcomes rather than surrogate metrics all need sustained attention. The methodology of connecting structural causal models to graph neural network architectures has been proposed. What remains is the work of building the validation infrastructure, earning clinical trust through transparent reporting of assumptions and limitations, and resisting the temptation to overstate what current methods can do. The goal is not to discover absolute causal truth in biological systems of extraordinary complexity but to construct causal approximations that improve the status quo. We can achieve models that degrade less under distribution shift, that separate therapeutic mechanisms from prescribing artefacts, and that give clinicians mechanistic reasoning they can interrogate rather than black-box predictions they have to accept on trust.

\bibliography{sn-bibliography}

\bmhead{Acknowledgements}

MM was supported by the Rhodes Trust and the EPSRC CDT Health Data Science. MB was supported by the German Academic Scholarship Foundation and the Clarendon Fund Scholarship. TZ was supported by the Royal Academy of Engineering under the Research Fellowship scheme.

\bmhead{Author contributions}

\bmhead{Competing interests}
We declare no competing interests.

\end{document}